\documentclass[12pt]{article}
\pdfoutput=1

\usepackage[a4paper, total={6in, 9in}]{geometry}
\usepackage[export]{adjustbox}
\usepackage{tikz}
\usepackage{amsmath,amsfonts, amssymb,amsthm,mathrsfs,bm,mathtools, graphicx, url, algorithm2e}
\usepackage{hyperref}

\newcommand{\cH}{\mathcal{H}}
\newtheorem{theorem}{Theorem}

\title{Support Collapse of Deep Gaussian Processes with Polynomial Kernels for a Wide Regime of Hyperparameters}
\author{Daryna Chernobrovkina and Steffen Gr\"unew\"alder \\
	University of York \\
Department of Mathematics \\
York, UK 
}
\date{}
\usepackage{times}
\usepackage{graphicx}

\DeclareMathOperator{\Var}{Var}

\begin{document}

\maketitle

\begin{abstract}%
 We analyze the prior that a Deep Gaussian Process with polynomial kernels induces. We observe that, even for relatively small depths, averaging effects occur within such a Deep Gaussian Process and that the prior can be analyzed and approximated effectively by means of the Berry-Esseen Theorem. 
 One of the key findings of this analysis is that, in the absence of careful hyper-parameter tuning, the prior of a Deep Gaussian Process either collapses rapidly towards zero as the depth increases or places negligible mass on low norm functions. This aligns well with experimental findings and mirrors known results for convolution based Deep Gaussian Processes. %
\end{abstract}


\section{Introduction}
Deep Gaussian processes (DGPs) have been introduced by  \cite{pmlr-v31-damianou13a} as a natural extension of Gaussian processes (GPs) that has been inspired by deep neural networks. Like deep neural networks, DGPs have multiple layers and each layer corresponds to an individual GP. It has recently been noted by \cite{NEURIPS2022_8c420176} that traditional GPs attain for certain compositional regression problems a strictly slower rate of convergence than the minimax optimal rate. This is demonstrated in \cite{NEURIPS2022_8c420176} by showing that for a class of generalized additive models any GP will be suboptimal, independently of the kernel function that is used. Generalized additive models can be regarded as a simple form of a compositional model with two layers. In contrast, \cite{JMLR:v24:21-0556} have shown that DGPs can attain for such problems the minimax optimal rate of convergence (up to logarithmic factors) when the DGPs are \textit{carefully tuned}. In fact, they show that DGPs are able to attain optimal rates of convergence for many compositional problems. Along similar lines, 
\cite{abraham2023deepgaussianprocesspriors} show that for nonlinear inverse problems DGPs can attain a rate of convergence that is polynomially faster than the rate that GPs with Mat\'ern kernel functions can attain when the unknown parameter has a compositional structure. One well known downside of DGPs is the difficulty of sampling from the posterior distribution. \cite{castillo2024} approach this problem by providing a particularly \textit{simple prior which facilitates posterior calculations} while guaranteeing adaptivity in the context of regression to both the smoothness of the unknown regression function and the compositional structure. 

In this paper,
we focus on the prior that a DGP places on a function space. We work in the context of polynomial kernels and we study the behavior of the priors as the depth of the DGP increases. We show that the prior is very sensitive to the hyperparameters that are used for the individual GPs and that small deviations of the `correct regime' of hyperparameters would either lead to an extremely tight concentration at zero or would result in prior measures that place negligible mass on functions of low norm. In earlier work, \cite{pmlr-v33-duvenaud14} have observed in experiments that the prior of a DGP with Gaussian kernels shows pathological behavior. They also analyzed the derivative of a DGP to get insight into this pathological behavior,  but did not provide an analysis of the behavior of the prior itself. \cite{DUN18} provide a deeper analysis by phrasing a DGP as a Markov chain and studying its ergodic behavior. In particular, \cite[Thm 4]{DUN18} states that
the output of a DGP becomes constant (in a form of point-wise convergence) as the depth increases when a Gaussian kernel is used and a condition on the parameters of the kernel is satisfied. They also study a DGP where instead of a composition of GPs a convolution of GPs is used. This form of a DGP differs from the DGPs that are commonly used in the literature \cite{pmlr-v31-damianou13a,pmlr-v33-duvenaud14,NEURIPS2022_8c420176, JMLR:v24:21-0556, castillo2024}, but  has the advantage that it is amenable to a convolution and Fourier theory based argument. This allows the authors to get deep insights into this type of DGPs. They find that 
for a convolutional DGP,  Fourier coefficients associated with the DGP converge either to zero or diverge (almost surely).  Furthermore, the eigenvalues of a covariance operator associated with the DGP control if the coefficients converge to zero or diverge \cite[Thm 16]{DUN18}.

One of the key research challenges in the area of DGPs  is to gain deeper insight into the behavior of standard DGPs. Such insight is  crucial to make sense of the `contradictory' observations in the literature: on the one hand, DGPs are often used successfully in practice \cite{pmlr-v31-damianou13a} and DGPs outperform GPs in a variety of statistical tasks in terms of rate of convergence \cite{JMLR:v24:21-0556} while, on the other hand, there is the pathological behavior of DGPs that has been observed in experiments and in convolutional DGPs \cite{pmlr-v33-duvenaud14,DUN18}. This research challenge is also far from trivial since the convolutional structure allows for  significant simplifications in the analysis of  \cite{DUN18}, and it is unclear how to get tight control of the behavior of a DGP in its absence.  

In the context of polynomial kernels, we develop an alternative approach that does not rely on convolutions and applies to standard DGPs. Our approach makes use of the fact that for polynomial kernel functions the sample paths of GPs lie within the reproducing kernel Hilbert space associated to that kernel function. Combining this fact with a Karhunen-Loève type decomposition of the GPs allows us to write the composition of GPs as a product of normally distributed vectors. We study these products then with the help of the Berry-Esseen Theorem. It is worth highlighting that earlier works focused on Law of Large Numbers and Ergodic type results which provide neither rates of convergence nor finite sample bounds. In contrast to that, the Berry-Esseen approach that we develop provides both rates of convergence and finite sample bounds. 

Our main result is Theorem \ref{the_thm}, which provides a bound on the approximation of a DGP $g_\ell \circ \ldots \circ g_1(x)$, where $g_1$ has covariance $k_1(x,y) = (xy + c)^{d_1}$, $c\geq 0$, and the $g_i$'s have covariance $k_i(x,y) = \sigma_i^2(xy)^{d_i}$, where $\sigma_i >0$ and the $d_i$'s  are non-zero integers. For such a process we find a normally distributed random variable $Y$ and a random sign $S$ such that 
\[
\sup_{x,t\in \mathbb{R}} |\Pr(g_\ell \circ \ldots \circ g_1(x) \leq t) - \Pr( S e^Y (g_1(x))^{c_1}  \leq t)| \leq 0.56 
\Bigl(\sum_{i=2}^\ell \sigma^2_{i,\log} \Bigr)^{-3/2} \sum_{i=2}^\ell \rho_{i,\log},
\]
where $\sigma^2_{\log},\rho_{\log}$ is the variance and absolute third moment of certain log-normal random variables. The constant $c_1$ is equal to  $d_2 + \ldots + d_\ell$ and will generally be very large. It is worth highlighting  that we have here an approximation of a DGP that consists of a product of a single GP, a random sign, and a log-normal random variable.  

Another important result that can be derived from the theorem is that when $\sigma_2=\ldots = \sigma_\ell =: \sigma$, then the median of the DGP converges rapidly to zero in $\ell$ if $\sigma < \exp((\gamma + \log 2)/2)$, where $\gamma$ is the Euler-Mascheroni constant, and diverges when $\sigma > \exp((\gamma + \log 2)/2)$. This is the same threshold that was found by \cite{DUN18} in the context of convolutional DGPs.

The remainder of the paper is organized as follows: in Section \ref{sec:prelim} we provide key definitions and results that we use throughout. In Section \ref{sec:prod_gauss}, we start with the simple case of products of Gaussian random variables; the motivation for this is that the main averaging effects that are at play are very transparent in this simple setting. Section  \ref{sec:MainSec}  is our main section. We start with the simple case of a product of GPs with linear kernels before approaching the case of polynomial kernels. 
In Section \ref{sec:discussion} we provide then a discussion of the results and we put these in perspective. In particular, we highlight challenges that need to be overcome to extend our results beyond the polynomial kernel case. There are also two appendices with technical results. In Appendix \ref{app:closed_form} we provide a variety of closed form expressions for moments of log-normal random variables that we use throughout, and Appendix \ref{app:DGPs} contains a variety of auxiliary results for DGPs that we use.

\subsection{Preliminaries}\label{sec:prelim}
A zero mean GP $g$  on $\mathbb{R}$ is a stochastic process which is fully specified by its covariance function $k(x,y), x,y \in \mathbb{R}$.  The covariance function $k$ is positive semi-definite.   The function $k$ is called the covariance function since $\Var(g(x)) = k(x,x)$ and $Cov(g(x) g(y)) = k(x,y)$ for all $x,y \in \mathbb{R}$.  In the context of kernel methods, one also calls $k$ the kernel function and we use the two terms interchangeably.  To each covariance function there corresponds a reproducing kernel Hilbert space (RKHS) $\cH_k$. In case that $\cH_k$ is finite dimensional it is known that the GP $g$ attains values in $\cH_k$.  In other words, the sample paths are RKHS functions when $\cH_k$ is finite dimensional. If $\cH_k$ is infinite dimensional then the sample paths lie almost surely not in $\cH_k$.  Often it is convenient to work with a so called feature map $\phi:\mathbb{R} \to \cH_k$ which satisfies $\langle \phi(x), \phi(y) \rangle =  k(x,y)$, where the inner product is here the inner product of $\cH_k$. In the finite dimensional case, we also write $ \phi(x)^\top \phi(y) = k(x,y)$.  A DGP  of depth $\ell$ on $\mathbb{R}$ is a composition of $\ell$ zero mean GPs  $g_\ell \circ \ldots \circ g_1$ with corresponding covariance functions $k_\ell, \ldots, k_1$.   

We make frequent use of the Central Limit Theorem (CLT) and different versions of the Berry-Esseen Theorem.  In particular, we use the following two versions of the Berry-Esseen Theorem, which guarantee uniform convergence of certain normalized sums to a Gaussian limit: (1) The first version that we use applies to zero mean i.i.d.  random variables $X_1,\ldots, X_n$ with variance $\Var(X_1) = \sigma^2$ and absolute third moment $\rho = E(|X_1|^3)$.  Let  $S_n = X_1 + \ldots + X_n$ then this version of the Berry-Esseen Theorem states that
\[
\sup_{x\in\mathbb{R}} |\Pr(n^{-1/2} \sigma^{-1} S_n \leq x) - \Phi(x) |  \leq \frac{0.336(\rho + 0.415\sigma^3)}{\sigma^3 n^{1/2}},
\] 
where $\Phi$ denotes the cumulative distribution (CDF) function of a standard normal random variable.

(2) The second version avoids the need of identically distributed random variables at the cost of a slightly more conflated theorem statement.  Consider again independent zero mean random variables $X_1,\ldots, X_n$ but now with individual variances $\Var(X_i) = \sigma_i^2$ and absolute third moments $\rho_i = E(|X_i|^3)$,  $i\leq n$. The second version of the Berry-Esseen Theorem states that
\[
\sup_{x\in\mathbb{R}} |\Pr\biggl( \frac{S_n}{\sqrt{\sigma_1^2 + \ldots + \sigma_n^2}} \leq x\biggr) - \Phi(x) |  \leq 0.56 
\Bigl(\sum_{i=1}^n \sigma_i^2\Bigr)^{-3/2} \sum_{i=1}^n \rho_i.
\] 
Note that the $n^{-1/2}$ factors that appear in the first version are subsumed in the variance terms; i.e.  when $\sigma_1 = \ldots = \sigma_n = \sigma$ then $\sqrt{\sigma_1^2 + \ldots + \sigma_n^2} = \sqrt{n} \sigma$ and when additionally $\rho_1 = \ldots = \rho_n = \rho$ then  $(\sum_{i=1}^n \sigma_i^2)^{-3/2} \sum_{i=1}^n \rho_i = \rho / \sqrt{n} \sigma^3$. 

Besides the definition of a DGP, all of the above results are classical and can be found in  textbooks such as \cite{kallenberg02}.

\section{Products of Gaussian Random Variables} \label{sec:prod_gauss}

\begin{figure}[t]
    \centering
    \begin{tabular}{ c @{\quad} c }
    \raisebox{-\height}{\includegraphics[valign=t,width=0.47\textwidth]{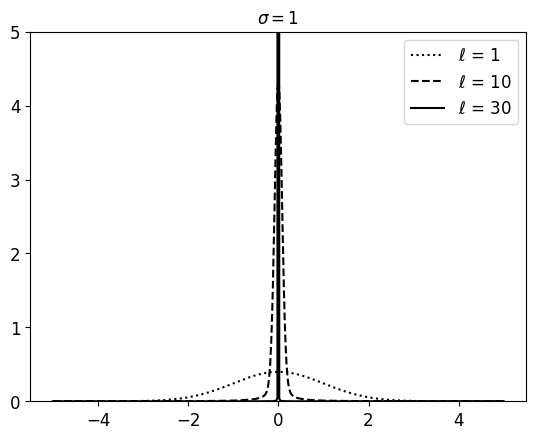}} &
   \raisebox{-\height - 0.33cm}{
   \begin{tikzpicture}
    \node[inner sep=0pt] (Pr) at (0,0)
    {\includegraphics[width=0.485\textwidth]{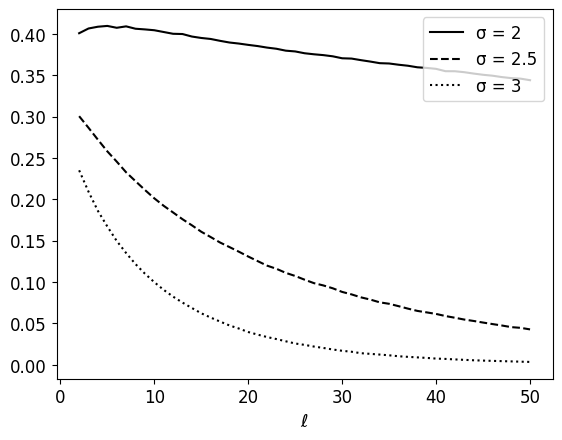}};
    \node[draw] at (1.2,0.6) {$\Pr(\prod_{i=1}^\ell |X_i| \leq 1/2)$};
    \end{tikzpicture} } \\
    \hspace{0.15cm} \small (a) & \hspace{0.5cm}
    \small (b)
  \end{tabular}
    \caption{(a) The densities of the product of $\ell = 1, 10, 30$ normally distributed random variables with mean $\mu = 0$ and variance $\sigma^2 = 1$ are shown. (b) The probability of the product attaining values around zero for larger $\sigma$ is shown.}
    \label{3_dens_prods}
\end{figure}

We start by analyzing the products of Gaussian random variables before approaching DGPs in the following section. We will see that such products are closely related to compositions of GPs with linear kernels. 
Let $X_1, ..., X_\ell$ be i.i.d. standard random variables with variance $\sigma^2>0$ and consider their product $\prod_{i = 1}^{\ell}X_i$. Figure \ref{3_dens_prods} plots the density of the product in dependence of $\ell$. Notice that the left plot uses $\sigma=1$ and that the density rapidly concentrates around zero in this case, as $\ell$ increases. The right plot considers larger values of $\sigma$ (the values are $2,2.5$ and $3$) and we can notice the opposite effect: the probability for the absolute value of the product to attain values below $1/2$ rapidly falls as $\ell$ increases. We will observe this effect repeatedly in other contexts.

\begin{figure}[t]
    \centering
    \hspace*{-0.3cm}
    \begin{tabular}{ c @{\quad} c }
    \raisebox{-\height}{\includegraphics[valign=t, width=0.49\textwidth]{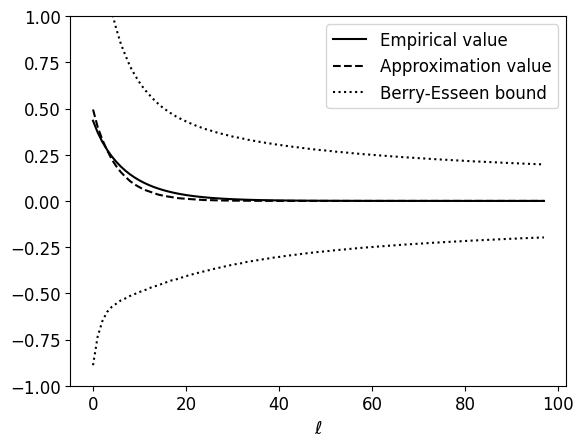}} &
    \raisebox{-\height}{\includegraphics[valign=t,width=0.49\textwidth]{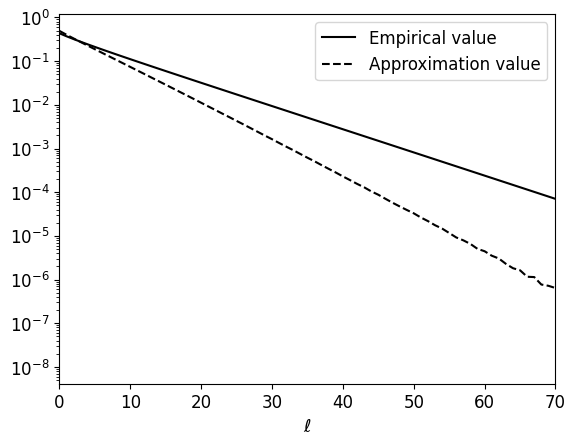}} \\
    \hspace{0.5cm} \small (a) & \hspace{0.4cm}
    \small (b)
    \end{tabular}
    \caption{(a) The probability of the scaled product and the log-normal approximation to attain values above $1/2$ are compared ($\sigma =1$). The plot is complemented by an error bound. (b) The same quantities are compared but on a logarithmic scale (the error bound is omitted).}
    \label{fig2}
\end{figure}

We will now aim to characterize the distribution of the product as $\ell$ increases. In order to do that, we apply the CLT to the product. We write the product as
$$\prod_{i = 1}^{\ell}X_i =  \left(\prod_{i = 1}^{\ell} S_i \right) \left(\prod_{i = 1}^{\ell} |X_i|\right),$$
where $S_i$ is the sign of $X_i$, 
\[S_i = \begin{cases}
    1 & \text{if } X_i \geq 0, \\
    -1 & \text{otherwise.}
\end{cases}
\]
Note that $S_i$ is independent of $|X_i|$ (Appendix \ref{sec:ind_S_absX}) and that $\prod_{i=1}^\ell S_i$ attains values $1$ and $-1$ each with probability $1/2$. If we take the logarithm of  $\prod_{i = 1}^{\ell} |X_i|$ then the CLT is applicable if the variance of
$\log |X_i|$ is finite. The variance of $\log|X_i|$ is, in fact, finite (See \eqref{eq:variance_log} in the Appendix) and the CLT can be applied. Under the assumption $X_1,\ldots, X_\ell$ are i.i.d. we can infer that 
\[\ell^{-1/2}\sum_{i=1}^{\ell} (\log|X_i| - E(\log|X_i|))  \overset{d}{\to} N(0, \Var(\log|X_1|)),
\]
where $d$ denotes convergence in distribution. In particular, for large $\ell$ the sum
$\ell^{-1/2}\sum_{i=1}^{\ell} \log|X_i|$  has approximately the distribution $N(\sqrt{\ell} E(\log|X_i|), \Var(\log|X_i|))$.
Furthermore, the continuous mapping theorem \cite[Thm 2.3]{vanderVaart1998} can be applied since the exponential function is continuous, and it follows that a normalized version of the product converges in distribution, 
\begin{align*} 
&\left(\prod_{i=1}^\ell \frac{|X_i|}{\exp(E(\log|X_i|))}\right)^{1/\sqrt{\ell}} 
= \exp\Bigl( \frac{1}{\sqrt{\ell}} \sum_{i=1}^{\ell} (\log|X_i| - E(\log|X_i|))\Bigr) \overset{d}{\to} e^{Z},
\end{align*}
where $Z$ is normally distributed with mean zero and variance $\Var(\log|X_1|)$. For large enough $\ell$ we then have the approximation,
\begin{align*}
\prod_{i=1}^\ell |X_i|^{1/\sqrt{\ell}} \approx e^{Z + \sqrt{\ell} E(\log|X_1|)} \text{\quad \quad (in distribution).}
\end{align*}
In other words, $\prod_{i=1}^\ell |X_i|^{1/\sqrt{\ell}}$ is approximately log normally distributed with mean parameter $\sqrt{\ell}E(\log|X_i|)$ and variance parameter $\Var(\log|X_i|)$. Figure \ref{fig2} shows a comparison of this approximation and the corresponding distribution of the scaled product (gained by sampling).

One might wonder if the normalizing factor $\ell^{-1/2}$ can be incorporated into the variance of the $X_i$ so that we can say something about the product $\prod_{i=1}^\ell \tilde X_i$ of suitably normalized Gaussian random variables $\tilde X_i$. This does not work, however, since $\Var(\log|X_i|) = \pi^2/8$ independently of the variance parameter $\sigma>0$. This is a consequence of $\Var(\log|aX_i|) = \Var( \log |a| + \log |X_i|) = \Var(\log|X_i|)$, which holds for any $a\in \mathbb{R}$.

\subsection{An Application of the Berry-Esseen Theorem} \label{sec:B-E_Application}
Ideally, we want to be able to infer properties of $\prod_{i=1}^\ell X_i$  or of $\prod_{i=1}^\ell X_i^\alpha$, $\alpha \in \mathbb{N}$. This will ultimately be useful for understanding how products of Gaussian processes with polynomial kernels behave. When following the earlier approach, we are led to expressions of the form $\sum_{i=1}^\ell \log |X_i|$ and $\alpha \sum_{i=1}^\ell \log |X_i|$. The leading $\alpha$ term in the latter expression is of minor importance. However, the lack of the normalizing factor $\ell^{-1/2}$ in front of the sums is a significant problem, since we cannot apply the CLT directly. This problem can be understood in terms of point-wise convergence. The CLT tells us that for any $x \in \mathbb{R}$, $\lim_{\ell\to \infty} \Pr(\ell^{-1/2} \sum_{i=1}^\ell ( \log |X_i| - E(\log|X_i|) ) \leq x) = \Phi(x/ \sigma_{\log}) $, where $\Phi$ denotes the CDF of a standard normal random variable
and $\sigma^2_{\log} = \Var(\log|X_1|) = \pi^2/8 $. To control the difference between the CDF of the unnormalized sum and $\Phi$ we can try
\[
\Pr\Bigl( \frac{1}{\sigma_{\log}} \sum_{i=1}^\ell (\log |X_i| - E(\log|X_i|)) \leq x\Bigr) 
= \Pr\Bigl(\frac{\ell^{-1/2}}{ \sigma_{\log}} \sum_{i=1}^\ell (\log |X_i| - E(\log|X_i|)) \leq \ell^{-1/2} x\Bigr)
\]
and hope that the latter expression gets close to $\Phi(\ell^{-1/2} x)$. However, the CLT does not allow us to infer this convergence since the location $\ell^{-1/2} x$ changes with $\ell$.

\begin{figure}[t]
    \centering
    \begin{tabular}{ c @{\quad} c } 
    \raisebox{-\height}{\includegraphics[valign=b,width=0.485\textwidth]{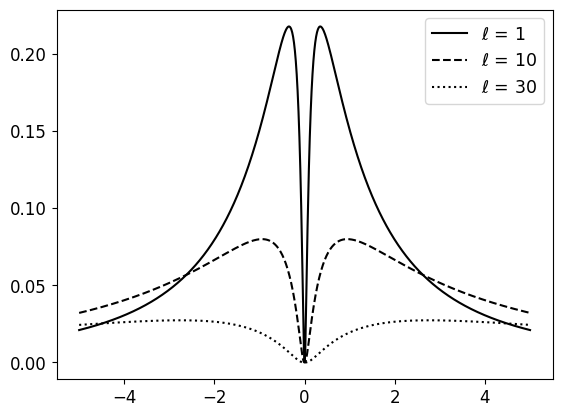}}&
    \raisebox{-\height-0.04cm}{
   \begin{tikzpicture}
    \node[inner sep=0pt] (Pr) at (0,0)
    {\includegraphics[valign=b,width=0.47\textwidth]{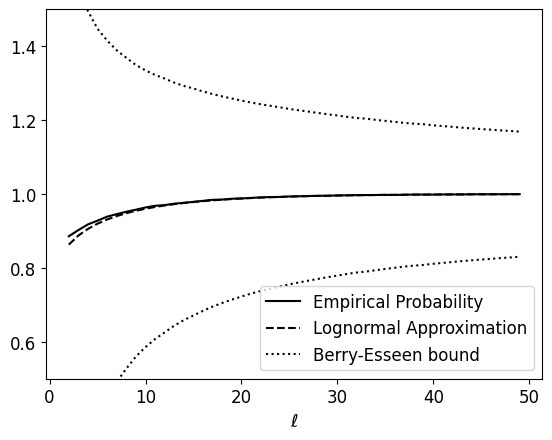}};
    \node[draw] at (1.2,2.2) {$\Pr(\prod_{i=1}^\ell |X_i| > 1/2)$};
    \draw[thick,loosely dashed,red] (-2.96,0.33) -- (3.45,0.33);
    \end{tikzpicture} } \\
    \hspace{0.45cm} \small (a) & \hspace{0.4cm}
    \small (b)
    \end{tabular}
    \caption{(a) The distribution of the product of $\ell = 1, 10$ and $30$ log-normal random variables with $\sigma=3$ is shown. (b) The probability for the product and the log-normal approximation to attain values above $1/2$ is shown ($\sigma = 3$).}
    \label{fig3}
\end{figure}

One way to address this nettle is to move from point-wise convergence to uniform convergence. This can be achieved by using the Berry-Esseen Theorem instead of the CLT. The Berry-Esseen Theorem guarantees that 
\[
\sup_{x\in \mathbb{R}} |  \Pr\Bigl(\ell^{-1/2} \sigma^{-1}_{\log}\sum_{i=1}^\ell (\log |X_i| - E(\log|X_i|)) \leq x\Bigr) - \Phi(x)| \leq  \frac{0.336 (\rho^\sigma_{\log} + 0.415 \sigma_{\log}^3) }{ \sqrt{\ell} \sigma_{\log}^3},
\]
where we assume that our Gaussian variables $X_1,\ldots, X_\ell$ are i.i.d., centered, and have variance $\sigma^2_{\log}$ and where we use the definition $\rho^\sigma_{\log} = E(|\log^3 |X_i||)$. We provide a closed-form expression of $\rho^\sigma_{\log}$ in Appendix \ref{app:closed_form} , \eqref{eq:rho_value}, as well as an easier to interpret bound \eqref{bnd:rho}. It is also easy to get very accurate approximations of $\rho^\sigma_{\log}$ through sampling. We can now infer that, uniformly in $x\in \mathbb{R}$,
\begin{align*}
&|\Pr\Bigl(\sigma_{\log}^{-1} \sum_{i=1}^\ell (\log |X_i| - E(\log|X_i|)) \leq x\Bigr) - \Phi(\ell^{-1/2} x) | \\
&= | \Pr\Bigl(\ell^{-1/2} \sigma_{\log}^{-1} \sum_{i=1}^\ell (\log |X_i| - E(\log|X_i|)) \leq \ell^{-1/2} x\Bigr) - \Phi(\ell^{-1/2} x)| \\
&\leq  \sup_{y\in \mathbb{R}} |  \Pr\Bigl(\ell^{-1/2} \sigma_{\log}^{-1}\sum_{i=1}^\ell (\log |X_i| - E(\log|X_i|)) \leq y\Bigr) - \Phi(y)| \\
&\leq \frac{0.336 (\rho^\sigma_{\log} + 0.415 \sigma_{\log}^3) }{ \sqrt{\ell} \sigma_{\log}^3}.
\end{align*}
We can rewrite this further to get an approximation of the law of $\sum_{i=1}^\ell \log |X_i|$,
\begin{align*}
\Pr\Bigl( \sum_{i=1}^\ell \log |X_i|   \leq \sigma_{\log} x + \ell E(\log|X_1|)\Bigr) &= \Pr\Bigl( \sigma_{\log}^{-1} \sum_{i=1}^\ell (\log |X_i| - E(\log|X_i|)) \leq x\Bigr) \\ &\approx \Phi(\ell^{-1/2} x).
\end{align*}
In other words, with $y = \sigma_{\log} x + \ell E(\log |X_1|)$,
\begin{equation} \label{eq:BE_ZApprox}
\Pr\Bigl( \sum_{i=1}^\ell \log |X_i|   \leq y \Bigr)
\approx \Phi(\ell^{-1/2} \sigma_{\log}^{-1} (y - \ell E(\log|X_1|))).
\end{equation}
If we let $Z \sim N(\ell E(\log|X_1|), \ell \sigma^2_{\log})$ then \eqref{eq:BE_ZApprox} implies 
\[
\sup_{x \in \mathbb{R}}| \Pr\Bigl( \sum_{i=1}^\ell \log |X_i|   \leq x \Bigr) - \Pr(Z \leq x)| \leq \frac{0.336 (\rho^\sigma_{\log} + 0.415 \sigma_{\log}^3) }{ \sqrt{\ell} \sigma_{\log}^3}.
\]
Since $\Pr(Z \leq x) = \Pr(e^Z \leq e^x)$, and similarly for $\sum_{i=1}^\ell \log|X_i|$, we find that
\begin{align}
\sup_{x \in \mathbb{R}}| \Pr\Bigl( \prod_{i=1}^\ell |X_i|   \leq x \Bigr) - \Pr(e^Z \leq x)| &= 
\sup_{x \in \mathbb{R}}| \Pr\Bigl( \prod_{i=1}^\ell |X_i|   \leq e^x \Bigr) - \Pr(e^Z \leq e^x)| \notag \\
&\leq \frac{0.336 (\rho^\sigma_{\log} + 0.415 \sigma_{\log}^3) }{ \sqrt{\ell} \sigma_{\log}^3}. \label{eq:BE_prod_approx}
\end{align}
The median of the log-normal random variable $e^Z$ is approximately $\exp(\ell(\log(\sigma) - 0.63))$. In particular, when $\sigma < e^{0.63} \approx 1.87$, the median approaches exponentially fast $0$, while when $\sigma > e^{0.63}$, the median diverges to infinity at an exponential rate in $\ell$. 
We demonstrate this divergence effect in Figure \ref{fig3} for $\prod_{i=1}^\ell X_i$ and the approximation $S_Z e^Z$, where $S_Z$ is a random variable that is independent of $Z$, and which attains values $+1$ and $-1$ with probability $1/2$ each.

\subsection{Convergence \& Approximation for Powers of $X$}
\label{sec:power_stdnorm}
We aim to generalize the above approach to products of the form $\prod_{i=1}^\ell X_i^\alpha$. When $\alpha$ is even then this product will always be positive and will be equal to $\prod_{i=1}^\ell |X_i|^\alpha$. When $\alpha$ is odd then 
$\prod_{i=1}^\ell X_i = S_\alpha \prod_{i=1}^\ell |X_i| $, where $S_\alpha$ attains  values $+1$ and $-1$ with equal probability. We will apply again the 
the Berry-Esseen Theorem to approximate the distribution of $\prod_{i=1}^{\ell} |X_i|^\alpha$. To this end, note that
\begin{align*}
&\sup_{x \in \mathbb{R}} |\Pr\Bigl(\ell^{-1/2} (\alpha \sigma_{\log})^{-1} \sum_{i=1}^{\ell} (\alpha \log |X_i| - \alpha E(\log |X_i|)) \leq x \Bigr) - \Phi(x) | \\
&\leq \frac{0.336(\alpha^3\rho^\sigma_{\log} + 0.415 (\alpha \sigma_{\log})^3)}{(\alpha \sigma_{\log})^3 \sqrt{\ell}},
\end{align*}
where the $\alpha$'s on the right side cancel. As in the previous section, 
after introducing the random variable $Z_\alpha \sim N(\ell \alpha E(\log|X_1|), \ell (\alpha \sigma_{\log})^2)$, we can infer that  
\begin{align*}
\sup_{x \in \mathbb{R}} |\Pr \Bigl(\prod_{i=1}^\ell |X_i|^{\alpha} \leq x\Bigr) - \Pr(e^{Z_{\alpha}} \leq x)| &= \sup_{x \in \mathbb{R}} |\Pr \Bigl(\prod_{i=1}^\ell |X_i|^{\alpha} \leq e^x\Bigr) - \Pr(e^{Z_{\alpha}} \leq e^x)| \\
&\leq \frac{0.336(\rho^\sigma_{\log} + 0.415  \sigma_{\log}^3)}{ \sigma_{\log}^3 \sqrt{\ell}}.    
\end{align*}
Note that $\alpha$ plays a similar role as $\ell$ and the distribution of $Z_\alpha$ for, say, $\alpha = 5$ and $\ell=10$ has the same mean as the random variable $Z_1$ with $\ell=50$. This implies that larger values of $\alpha$ lead to a more rapid collapse of the support of $e^{Z_\alpha}$. For example,
the median of the variable $e^{Z_\alpha}$ is $\ell \alpha (\log(\sigma) - 0.63)$. As the factor $\alpha$ increases, we have an even faster convergence of the median to $0$ for $\sigma < 1.87$ and divergence to infinity when $\sigma > 1.88$.

\begin{figure}[ht]
    \centering
    \hspace*{-0.3cm}
    \begin{tabular}{ c @{\quad} c }
    \includegraphics[width=2.85in]{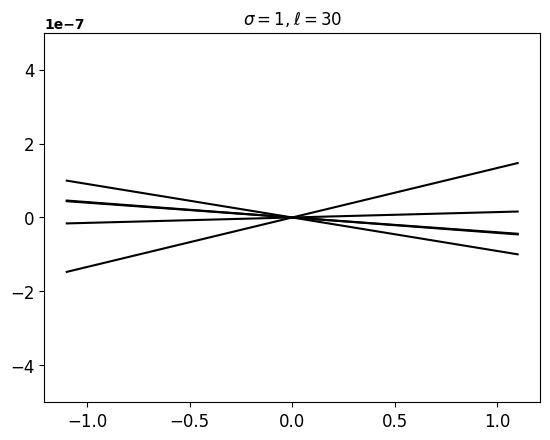}&
    \includegraphics[width=2.95in]{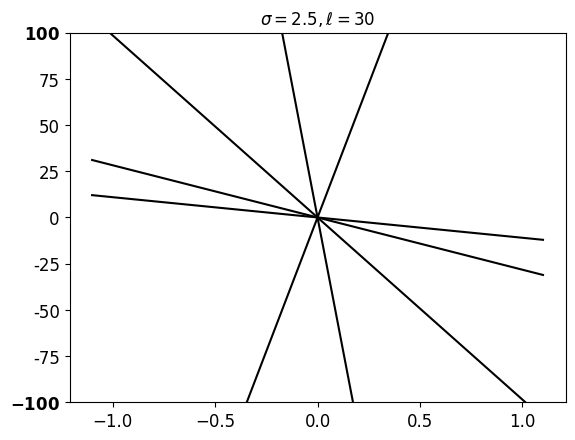} \\
    \hspace{0.3cm} \small (a)  & \hspace{0.6cm}
    \small (b)
    \end{tabular}
    \caption{(a) Five draws from a DGP with $\ell=30$ layers, a linear kernel, and $\sigma =1$ is shown. Notice the scale of the $y$-axis. (b) As for (a) but with $\sigma=2.5$.}
    \label{fig4}
\end{figure}

\section{Limit Distributions of Deep Gaussian Processes} \label{sec:MainSec}
The results in the last section on the distribution of $\prod_{i=1}^\ell X_i$, where $X_i$ are normally distributed, have natural links to the distribution of deep GPs. To see this, let us start with the simple case where the kernel functions of the different layers are all $k(x,y) = \sigma^2 xy$ and $x,y \in \mathbb{R}$. Now, consider  $\ell$ independent Gaussian processes $g_1,\ldots, g_\ell$ each of which has zero mean and kernel function $k$. The process $g_\ell \circ \ldots \circ g_1$ is a Deep GP on $\mathbb{R}$. The kernel  $k(x,y)$ can also be written as an inner product of a feature map $\phi$. In particular, if we use $\phi(x) = \sigma x$ then $k(x,y) = \phi(x) \phi(y)$. Each of the individual GPs attains values in the RKHS  $\mathcal{H}_k$ that corresponds to $k$. Note that this RKHS can
be written as $\{h: \mathbb{R} \to \mathbb{R} : h(x) = \alpha \phi(y) \phi(x), \alpha, y\in \mathbb{R} \}
= \{h: \mathbb{R} \to \mathbb{R} : h(x) = \alpha \phi(1) \phi(x), y\in \mathbb{R} \}$ since $\phi$ is linear in this context. Since $\phi(1) = \sigma$ we know that each path drawn from the GP will be of the form $\alpha \sigma \phi(x)$, where the slope $\alpha$ changes depending on our draw from the GP. Let us use the notation $\alpha^{(i)}_\omega$ to denote the slope corresponding to the draw from GP $g_i$ for experiment $\omega \in \Omega$, i.e. $\alpha_\omega^{(i)}$ is the random variable that corresponds to the slope of the paths drawn from $g_i$. Let us now take a look at $g_1$ and $\alpha_\omega^{(1)}$. Fixing some $x\in \mathbb{R}$, we know that $g_1(x)$ is a zero mean Gaussian random variable with variance $k(x,x) = \sigma^2 x^2$. We also know that $g_1(x) = \alpha_\omega^{(1)} \sigma \phi(x) = \alpha_\omega^{(1)} \sigma^2 x$ and $\alpha_\omega^{(1)} = g_1(x) / \sigma^2 x$. In other words, $\alpha_\omega^{(1)}$ is a Gaussian random variable with zero mean and variance $1/\sigma^2$. Hence, with $U_1 = \sigma^2 \alpha_\omega^{(1)} \sim N(0,\sigma^2)$ it follows that
$g_1(x) = \alpha_\omega^{(1)} \sigma \phi(x) = U_1 x$.
By induction we can generalize this to 
\[
g_\ell \circ \ldots \circ g_1(x) = \prod_{i=1}^\ell U_i x,
\]
where $U_1,\ldots, U_\ell$ are i.i.d. with distribution  $N(0,\sigma^2)$. The independence of $U_1,\ldots, U_\ell$ follows right away from the independence of $g_1,\ldots, g_\ell$ since each $U_i$ is a function of $g_i$.

\subsection{Approximation of DGPs with Linear Kernels}
From the previous section, we know that $g_\ell \circ \ldots \circ g_1(x)$ can be written as $x S \prod_{i=1}^\ell |U_i|$, where $S$ is independent of the $U_i$'s and attains values $+1$ and $-1$ with equal probability. The statement about $S$ follows by the same argument as in Section \ref{sec:prod_gauss}. We can now approximate the distribution of the product by means of the Berry-Esseen Theorem.  In particular, for large $\ell$ the product $\prod_{i=1}^\ell |U_i|$ will  be close in distribution to  $e^Z$, where $Z \sim N(\ell E(\log |U_1|), \ell \sigma_{\log}^2)$, with $\sigma_{\log}^2 = \Var(\log |U_1|)$ and $\rho_{\log}^\sigma = E(|\log|U_1||^3)$. In particular, by following the same argument as in Section \ref{sec:prod_gauss}, and by incorporating the random signs and $x$, we find that
\begin{align*}
\sup_{x,c \in \mathbb{R}} \Bigl|\Pr \Bigl(g_\ell \circ \ldots g_1(x) \leq c \Bigr) - \Pr( S x e^{Z}  \leq c)\Bigr| &= 
\sup_{x,c \in \mathbb{R}} \Bigl|\Pr \Bigl(S x \prod_{i=1}^\ell |U_i|  \leq c \Bigr) - \Pr(S x e^{Z}  \leq c)\Bigr| \\ 
& \leq \frac{0.336(\rho_{\log}^\sigma + 0.415 \sigma_{\log}^3)}{\sqrt{\ell}\sigma_{\log}^3}.
\end{align*}
It is worth highlighting that this bound holds uniformly over all values $x$. In fact, in the linear case, this follows right away since the Berry-Esseen bound is uniform in $c$ and we can use a simple substitution from $c/x$ to $c$ to infer that the bound also holds  uniformly in $x$.

\subsection{Approximation of DGPs with Polynomial Kernels} \label{sec:DGPApprox}
We will now extend the above results to DGPs of the form $g_\ell \circ \ldots \circ g_1$, where the GP $g_1$ 
has a polynomial kernel of order $d_1$, that is $k_1(x,y) = (xy + c)^{d_1}$, where $d_1 >0$ is some integer and $c \geq 0$, and the successive GPs have kernels $k_i(x,y) = \sigma_i^2 (xy)^{d_i}$, $d_i \geq 1, \sigma_i > 0$ and $i\geq 2$.  
The GP $g_1$ can be written in the following way (see Appendix \ref{app:polynomialKernel} on p. \pageref{app:polynomialKernel}),
$$
g_1(x) = \sum_{i=1}^{d+1} Z_i \phi_i(x)
= \sum_{i=0}^{d} \binom{d}{i}^{1/2} Z_{i+1} x^{d-i} c^{i/2},
$$
where $(Z_1,\ldots, Z_{d+1})^\top \sim N(0, I)$ and $\phi_i(x) = \binom{d}{i-1}^{1/2} x^{d-i+1} c^{(i-1)/2}$, $i\leq d+1$. Similarly, there are independent random variables
$Y_i \sim N(0, \sigma_i^2)$, which are also independent of $Z_1,\ldots,Z_{d+1}$, and such that
$g_i(x) = Y_i x^{d_i}$, for all $2\leq i \leq \ell$. We can therefore write the DGP as 
\begin{align*}
g_\ell \circ \dots \circ g_1(x) &= Y_\ell 
(Y_{\ell-1})^{d^\downarrow_1} (Y_{\ell -2})^{d^\downarrow_2} \times \dots \times (Y_{2})^{d^\downarrow_{\ell-2}} \Bigl(\sum_{i=1}^{d+1}Z_i \phi_i (x) \Bigr)^{d^\downarrow_{\ell-1}}, 
\end{align*}
where we use the notation $d^\downarrow_i = \sum_{j=0}^{i-1} d_{\ell-j}$, for $i=1, \dots \ell-1$. Taking the logarithm of the product of the absolute values of the $Y$-terms gives us $d^\downarrow_{\ell-2} \log|Y_2| + \dots + d^\downarrow_{1} \log|Y_{\ell-1}| 
+  \log|Y_{\ell}| =  \sum_{j=2}^{\ell}c_j \log|Y_j|$,
where $c_j = d^\downarrow_{\ell-j}$ for $j=2, \dots \ell-1$, and $c_\ell =1$. We are now in a position to apply the Berry-Esseen Theorem for \textit{non-identically distributed} random variables. To set this up, let $\sigma^2_{i,\log} = c_i^2 \Var(\log|Y_i|)$ and $\rho_{i,\log} =
c_i^3 E(| \log|Y_i||^3)$ for $2 \leq i\leq \ell$. Then
\begin{align} \label{eq:BE_forPoly}
&\sup_{x\in \mathbb{R}} |\Pr((\sigma^2_{2,\log} + \dots + \sigma^2_{\ell,\log})^{-1/2} \sum_{j=2}^\ell (c_j \log |Y_j| - c_jE(\log|Y_j|))\leq x) - \Phi(x)| \notag \\ 
&\leq 0.56 
\Bigl(\sum_{i=1}^n \sigma_{i,\log}^2\Bigr)^{-3/2} \sum_{i=1}^n \rho_{i,\log}.
\end{align}
As earlier, we can translate this statement  into a statement about $\sum_{j=2}^\ell c_j \log|Y_j|$ by means of substitution. For a given $x$, let $y = (\sigma^2_{2,\log}+\dots+\sigma^2_{\ell,\log})^{1/2} x + \sum_{j=2}^\ell c_jE(\log|Y_j|)$, then  
\[
|\Pr\Bigl(\sum_{j=2}^\ell c_j \log |Y_j| \leq y\Bigr) - \Phi\Bigl((\sigma^2_{2,\log}+\dots+\sigma^2_{\ell,\log})^{-1/2}\bigl(y- \sum_{j=2}^\ell c_jE(\log|Y_j|)\bigr)\Bigr)|
\]
is also upper bounded by the right side of \eqref{eq:BE_forPoly}. In fact, this bound holds uniformly over all $y\in\mathbb{R}$. Let us introduce a random variable $Y$ that is independent of $Z_1,\ldots, Z_{d+1}, S_2,\ldots, S_\ell$ and which has the law $N(\sum_{i=2}^\ell c_iE(\log|Y_i|), 
\sum_{i=2}^\ell \sigma^2_{i,\log})$, then
\begin{align*}
\sup_{y\in \mathbb{R}} |\Pr\Bigl( |Y_\ell| \prod_{i=2}^{\ell-1} |Y_i|^{c_i} \leq y\Bigr)
- \Pr(e^Y \leq y)| = \sup_{y\in \mathbb{R}} |\Pr\Bigl( \sum_{j=2}^\ell c_j \log |Y_j| \leq y \Bigr) - \Pr(Y \leq y)| 
\end{align*}
and the latter term is again upper bounded by the right side of \eqref{eq:BE_forPoly}.
We can translate this into a statement about the product of the $Y_i$'s by observing that  $Y_\ell \prod_{i=2}^{\ell-1} Y_i^{c_i} = S_\ell \prod_{i\in I} S_i  |Y_\ell| \prod_{i=2}^{\ell-1} |Y_i|^{c_i}$, where $I \subset \{2,\ldots, \ell\}$ are the indices which correspond to odd  $c_i$ values and $S_i$ is the sign of $Y_i$ for all $i\leq \ell$. Since the different $S_i$'s are independent of each other and independent of the $|Y_i|$'s it follows that $S = S_\ell \prod_{i\in I} S_i$ is a random variable that is independent of the $|Y_i|$'s (in fact, $S$ is also independent of $Z_1,\ldots, Z_{d+1}$) and it attains values $+1$ and $-1$ with equal probability. There is one final technical hurdle in the way to an approximation of the DGP. We need to multiply both the product and the approximation by the random variable $(g_1(x))^{c_1}$ (strictly speaking, we can have two different probability spaces and the new space needs to contain a copy of $(g_1(x))^{c_1}$). In any case, we can relate the two distributions that include $(g_1(x))^{c_1}$ by means of  a conditional expectation argument, which we provide in Appendix \ref{sec:conditioning}. 
From here we get directly to an approximation of the DGP. We summarize this statement in the following theorem.
\begin{theorem} \label{the_thm}
Given a DGP $g_\ell \circ \ldots \circ g_1$ on $\mathbb{R}$ with $\ell$-layers and corresponding independent GPs $g_1, \ldots, g_\ell$ with covariance functions $k_1(x,y) = (xy + c)^{d_1}$, $c\geq 0$, and $k_i(x,y) = \sigma_i^2(xy)^{d_i}$ where $\sigma_i >0$, $ 2 \leq i\leq \ell$, and $d_1,\ldots, d_\ell \geq 1$ are  integers. There exist independent $Y_2,\ldots, Y_\ell$, such that each $Y_i$ has  distribution $N(0,\sigma_i^2)$ and $g_i(x) = Y_i x^{d_i}$. 
For $2 \leq i \leq \ell$, 
let $\sigma^2_{i,\log} = c_i^2 \Var(\log|Y_i|)$ and $\rho_{i,\log} = c_i^3 E(|\log|Y_i||^3)$.
We have the following approximation of the DGP:  
\[
\sup_{x,t\in \mathbb{R}} |\Pr(g_\ell \circ \ldots \circ g_1(x) \leq t) - \Pr( S e^Y (g_1(x))^{c_1}  \leq t)| \leq 0.56 
\Bigl(\sum_{i=2}^\ell \sigma^2_{i,\log} \Bigr)^{-3/2} \sum_{i=2}^\ell \rho_{i,\log},
\]
where 
\[Y \sim N\Bigl(\sum_{i=2}^\ell c_i E(\log|Y_i|), \sum_{i=2}^\ell c_i^2 \Var(\log|Y_i|)\Bigr),\] 
and $c_\ell = 1$, $c_i = \sum_{j=i+1}^\ell d_j$, for $1 \leq i \leq \ell-1$. The random sign $S$ attains values $+1$ and $-1$  with equal probability. Furthermore, $g_1$, $S$ and $Y$ are independent and we can write
\[
g_1(x) = \sum_{i=0}^{d_1} \binom{d_1}{i}^{1/2} Z_{i+1} x^{d_1-i} c^{i/2},
\]
with $Z_1,\ldots, Z_n$  independent standard normal random variables that are independent of $S$ and $Y$.
\end{theorem}

\paragraph{Example ($d_2 = \ldots = d_\ell = 2$):} It is instructive to analyze the distribution of $S (g_1(x))^{c_1} e^Y$ and the Berry-Esseen bound in a concrete setting. Assume that $d_2 = \ldots = d_\ell = 2$ and $\sigma_2 = \ldots = \sigma_\ell = \sigma$, for some $\sigma$ that we will vary, and let $\ell \geq 2$. We show in Appendix \ref{app:BE_bnd_d2} (Eq. \eqref{eq:gather_BE}) that the \textit{Berry Esseen bound} takes in this setting the form 
\begin{align*}
0.56 \Bigl(\sum_{i=2}^\ell \sigma^2_{i,\log} \Bigr)^{-3/2} \sum_{i=2}^\ell \rho_{i,\log} \leq 3 \ell^{-1/2} \frac{E(|\log|Y_1||^3)}{(\Var(\log|Y_1|))^{3/2}}.
\end{align*}
Note that the bound improves with the familiar $\ell^{-1/2}$ rate. In terms of the dependence on $\sigma$, recall that $\Var(\log|Y_1|)$ is independent of $\sigma$ and the bound in Appendix \ref{pg:bnd_on_rho} on p. \pageref{pg:bnd_on_rho} shows that the dependence of $E(|\log|Y_1||^3)$ on $\sigma$ is at most logarithmical.

In terms of the distribution of $Y$, first note that for $2\leq i \leq \ell -1$ the coefficients become $c_i = 2 (\ell - i -1)$ and the \textit{mean} of $Y$ becomes (App. \ref{app:BE_bnd_d2}, Eq. \eqref{eq:gather_Mean}),
\[
E(\sum_{i=2}^\ell c_i E(\log|Y_i|))  = (\ell(\ell-1)-1) E(\log|Y_i|))  \approx  ((\ell-1)^2 + \ell)
 (\log(\sigma) -0.63).
\]
Similarly, the \textit{variance} becomes (App \ref{app:BE_bnd_d2}, Eq. \eqref{eq:gather_Var})
\[
\sum_{i=2}^\ell c_i^2 \Var(\log|Y_i|) = \Bigl(\frac{2\ell(\ell-1)(2 \ell - 1)}{3} -3\Bigr) \Var(\log|Y_1|) = \frac{\pi^2}{8} \Bigl(\frac{2\ell(\ell-1)(2 \ell - 1)}{3} -3\Bigr).
\]
In terms of the log-normal random variable $e^Y$ we have a similar effect as in the earlier settings: the \textit{median of $e^Y$} is $\exp( (1/2) ((\ell-1)^2 + \ell)(\log(\sigma) -(\gamma + \log 2)/2) )$ which approaches rapidly zero when $\sigma < \exp((\gamma + \log 2)/2) \approx 1.88$ and, otherwise, diverges to infinity as $\ell$ increases.
In terms of the approximation $S e^Y (g_1(x))^{c_1}$ of the DGP, note that $c_1 = 2(\ell-1)$ and 
\[
(g_1(x))^{c_1} = \biggl(\sum_{i=0}^{d_1} \binom{d_1}{i}^{1/2} Z_{i+1} x^{d_1-i} c^{i/2} \biggr)^{2(\ell-1)}.
\]
Hence, we have two terms of large order that interact multiplicatively. If $\sigma < 1.88$ and the sum $\sum_{i=0}^{d_1} \binom{d_1}{i}^{1/2} Z_{i+1} x^{d_1-i} c^{i/2}$ is strictly smaller than one then we expect the DGP to attain tiny values.  Similarly, when $\sigma > 1.88$ and the sum attains value above one then we expect the DGP to attain huge values. The most interesting case is the case where the two multiplicative terms compete: for simplicity consider the median value of $e^Y$, let $\sigma =1$, and assume that the sum attains a value of $a> 1$, then the product of these two terms is approximately of the order
$e^{-\ell^2} \times a^{2 \ell}$
and the term $e^Y$ will dominate when $\ell$ growth. The probability for attaining large values depends obviously on $x$; the larger $|x|$ the higher the probability to observe large values $a$ and the longer it will take for $e^{-\ell^2}$ to dominate.

\section{Discussion} \label{sec:discussion}
In this paper, we demonstrated that, for a range of polynomial kernels, a DGP will either concentrate around zero or it will generate functions that have increasingly large norms. The regime where these two pathologies do not occur becomes increasingly smaller as the size of the DGP increases. One can also note that kernels with higher order polynomials will likely give rise to DGPs that show this pathological behavior earlier (we discuss this phenomenon in the context of products of normally distributed random variables in Section \ref{sec:power_stdnorm}). Another important observation is that these effects arise due to averaging effects across the layers. One can break these averaging effects in the context of polynomial kernels by choosing the degree of some kernels significantly higher than of others. Then only a few GPs will dominate. This, however, will be similar to a DGP with fewer layers, which defeats the purpose of DGPs.

There is a wide range of open questions, that are important to address. One of the most important questions is how to generalize our approach to a larger class of kernel functions. A major hurdle that needs to be overcome to be able to generalize the results is that GPs with infinite dimensional RKHSs do not attain values within their RKHS. In all likelihood, the fact that the RKHS lies  dense in the space of functions in which the GP attains values will be crucial for extensions. 
Another important question is why DGPs show near optimal asymptotic rates of convergence in various problems given their pathological behavior. Is this simply the Bernstein-von-Mises Theorem at work, or is there a deeper reason?

\newpage
\appendix
\section{Closed-Form Solutions \& Bounds} \label{app:closed_form}

\subsection{Closed form expressions for $E(\log |X|)$,  $E(\log^2|X|)$,  $\Var(\log|X|)$ and $E(|\log^3 |X||)$}
It is known that the values $E(\log |X|)$ and $E(\log^2|X|)$ can be derived analytically when $X$ is normally distributed. With some more work one can also derive a closed-form expression for $E(|\log^3|X||)$ whenever the variance $\sigma^2$ of $X$ satisfies $\sigma^2 > 1/2$. We complement this last result with the simple bound $E(|\log^3|X||) \leq (E(\log^4|X|))^{3/4}$ for the case where $\sigma^2 \leq 1/2$.
For the reader's convenience, we present the various derivations in detail below. 

\paragraph{Applications of the $\Gamma$ function and the incomplete $\Gamma$ functions.}
The first step is to derive
 closed-form expressions for $\int_0^\infty e^{-x^2} \log |x| \, dx, \int_0^\infty e^{-x^2} \log^2 |x| \, dx$ and
$\int_0^\infty e^{-x^2} |\log^3 |x|| \, dx$. An easy argument to derive these uses the $\Gamma$ function
\begin{equation} \label{def:Gamma}
\Gamma(s) = \int_0^\infty x^{s-1} e^{-x}  \,dx,
\end{equation}
as well as the  upper incomplete $\Gamma$ function
\[
\Gamma(s,x) = \int_{x}^\infty t^{s-1} e^{-t} \,dt,
\]
and the lower incomplete $\Gamma$ function
\[
\gamma(s,x) = \int_{0}^x t^{s-1} e^{-t} \,dt,
\]
where for us it suffices to consider real valued $s > 0$, and where $x > 0$. 

Taking derivatives on both sides of \eqref{def:Gamma}, where we can exchange the derivative and the integral since the integral is defined for all $s > 0$, the derivative with respect to $s$ exists in a neighborhood around $s=1/2$, and there exists an integrable function that bounds the absolute value of the derivative from above (\cite{fremlin_v1}, Th.123D). Moreover, using the di-gamma function $\psi$,  which has the property that $\psi(s) = \Gamma'(s) / \Gamma(s)$, for all $s > 0$, gives
\[
\Gamma(s) \psi(s) =  \int_0^\infty x^{s-1} e^{-x} \log( x )  \,dx.
\]
In particular, for $s=1/2$,
\[
\Gamma(1/2) \psi(1/2) = \int_0^\infty x^{-1/2} e^{-x} \log( x )  \,dx = 2 \int_0^\infty e^{-x^2} \log( x^2) \, dx = 4 \int_0^\infty e^{-x^2} \log(x) \, dx
\]
and 
\begin{equation}
\int_0^\infty e^{-x^2} \log(x) \, dx = \frac{\Gamma(1/2) \psi(1/2)}{4} = - \frac{\sqrt{\pi}(\gamma + 2 \log 2 )}{4} =  \frac{\Gamma'(1/2)}{4}. \label{eq:gamma_1}
\end{equation}
Similarly, we can approach the integral for $\log^2|x|$. Analogously to the calculation above, we can exchange the derivative and the integral by (\cite{fremlin_v1}, Th.123D). Taking the second derivative of the $\Gamma$ function, yields
\begin{align*}
\frac{d}{ds} \Gamma(s) \psi(s) = \int_0^\infty x^{s-1} e^{-x} \log^2( x )  \,dx.
\end{align*}
On the left we have
\[
\frac{d}{ds} \Gamma(s) \psi(s) = \Gamma'(s) \psi(s) + \Gamma(s) \psi'(s) = \Gamma(s) \psi^2(s) + \Gamma(s) \psi'(s).
\]
The derivative $\psi'$ can be found under the name polygamma function. Evaluating the integral at $s=1/2$ yields
\[
2 \int_0^\infty (1/2) x^{-1/2} e^{-x} \log^2( x )  \,dx
= 2 \int_0^\infty  e^{-x^2} \log^2( x^2 )  \,dx
= 8 \int_0^\infty  e^{-x^2} \log^2( x )  \,dx
\]
and
\[
\int_0^\infty  e^{-x^2} \log^2( x )  \,dx =
\frac{\Gamma(1/2) \psi^2(1/2) + \Gamma(1/2) \psi'(1/2)}{8}
= \frac{\sqrt{\pi}( (2 \log 2 + \gamma)^2 + 3 \zeta(2) )}{8},
\]
where $\zeta$ is the Riemman-Zeta function.
It is known that $\zeta(2) = \pi^2/6$. For later use we note that
\begin{equation}
\Gamma''(1/2) = \sqrt{\pi}( (2 \log 2 + \gamma)^2 + 3 \zeta(2) ). \label{eq:gamma_2}
\end{equation}

Finding an analytic solution for $\rho = E(|\log^3|X||)$ when $\sigma^2 > 1/2$ requires the use of the incomplete $\Gamma$ functions. With the same arguments as for the $\Gamma$ function, one can show that:
\begin{gather*}
\frac{d}{d s} \Gamma(s,1/\sqrt{2}\sigma) \Big|_{s=1/2} = 4 \int_{1/\sqrt{2}\sigma}^\infty \log(x)  e^{-x^2} \, dx,  \\
\frac{d}{d s} \gamma(s,1/\sqrt{2}\sigma) \Big|_{s=1/2} = 4 \int_0^{1/\sqrt{2}\sigma} \log(x) e^{-x^2} \, dx, \\
\frac{d^2}{d s^2} \Gamma(s,1/\sqrt{2}\sigma) \Big|_{s=1/2} = 8 \int_{1/\sqrt{2}\sigma}^\infty \log^2(x)  e^{-x^2} \, dx,  \\
\frac{d^2}{d s^2} \gamma(s,1/\sqrt{2}\sigma) \Big|_{s=1/2} = 8 \int_0^{1/\sqrt{2}\sigma} \log^2(x) e^{-x^2} \, dx, \\
\frac{d^3}{d s^3} \Gamma(s,1/\sqrt{2}\sigma) \Big|_{s=1/2} = 16 \int_{1/\sqrt{2}\sigma}^\infty \log^3(x)  e^{-x^2} \, dx,  \\
\frac{d^3}{d s^3} \gamma(s,1/\sqrt{2}\sigma) \Big|_{s=1/2} = 16 \int_0^{1/\sqrt{2}\sigma} \log^3(x) e^{-x^2} \, dx.
\end{gather*}
The challenge is to find closed-form expressions for the derivatives of the incomplete $\Gamma$ functions. Because, 
$\Gamma(s) = \Gamma(s,1/\sqrt{2}\sigma) + \gamma(s,1/\sqrt{2}\sigma)$ it suffices to find the derivatives of $\Gamma(s)$ and $\Gamma(s,1/\sqrt{2}\sigma)$. We calculated the first and second derivatives of $\Gamma(s)$ already and the third and the fourth derivatives are also not hard to derive: 
\begin{align}
 \frac{d^3}{ds^3} \Gamma(s) \Big|_{s=1/2}  &= \frac{d}{ds}(\Gamma(s) \psi^2(s) + \Gamma(s) \psi'(s)) \Big|_{s=1/2} = 
    \Gamma(s) (\psi^3(s) + 3\psi(s) \psi'(s) + \psi''(s)) \Big|_{s=1/2}  \notag\\
    &=-\sqrt{\pi}((2 \log2 + \gamma)^3 + 9(2\log2 + \gamma)\zeta(2) + 14 \zeta(3)). \label{eq:gamma_3}
\end{align}

\begin{align}
\frac{d^4}{ds^4} \Gamma(s) \Big|_{s=1/2}  &= \frac{d}{ds}(\Gamma(s) (\psi^3(s) + 3\psi(s) \psi'(s) + \psi''(s)) )\Big|_{s=1/2} \notag\\ 
&= \Gamma(s) (\psi^4(s) + 6 \psi'(s) \psi^2(s) + \psi'''(s) +  4 \psi''(s) \psi(s) + 3(\psi'(s))^2)\Big|_{s=1/2} \notag\\
&= \sqrt{\pi}((2\log2 + \gamma)^4 + 18(2\log2 + \gamma)^2 \zeta(2) \notag \\
& \quad\quad\quad\enspace + 56(2\log2 + \gamma)\zeta(3) + 27\zeta^2(2) + 90 \zeta(4)). \label{eq:gamma_4}
\end{align}
The derivatives of $\Gamma(s,1/\sqrt{2}\sigma)$ are more challenging to derive and we are using results from \cite[p.156/157]{GED90} which apply when $\sigma^2 > 1/2$. 
The first derivative is
\begin{equation}
\frac{d}{d s} \Gamma(s,1/\sqrt{2}\sigma) \Big|_{s=1/2} = 
-\log(\sqrt{2} \sigma) \Gamma(1/2,1/\sqrt{2} \sigma) + \frac{1}{\sqrt{2} \sigma} T(3,1/2,1/\sqrt{2} \sigma), \label{eq:in_gamma_1}
\end{equation}
where the function $T$ is defined in \cite{GED90}. Further below, we derive the particular values of $T$ that we need. 
The second derivative is 
\begin{align}
\frac{d^2}{d s^2} \Gamma\left(s,\frac{1}{\sqrt{2}\sigma}\right) \Big|_{s=1/2} =& 
\log^2(\sqrt{2} \sigma) \Gamma\left(\frac{1}{2},\frac{1}{\sqrt{2} \sigma}\right) \notag \\
&+ \frac{2}{\sqrt{2} \sigma}\Bigl( - \log(\sqrt{2} \sigma)   T\left(3,\frac{1}{2},\frac{1}{\sqrt{2} \sigma}\right) + T\left(4,\frac{1}{2},\frac{1}{\sqrt{2} \sigma}\right) \Bigr),
\label{eq:in_gamma_2}
\end{align}
and the third derivative is
\begin{align}
\frac{d^3}{d s^3} \Gamma\left(s,\frac{1}{
\sqrt{2} \sigma}\right) \Big|_{s=1/2} 
= &-\log^3(\sqrt{2} \sigma) \Gamma\left(s,\frac{1}{
\sqrt{2} \sigma} \right)
+ \frac{3}{\sqrt{2}\sigma} 
\biggl( \log^2(\sqrt{2} \sigma) T\left(3,\frac{1}{2},\frac{1}{\sqrt{2} \sigma}\right) \notag \\
&\quad \quad \quad - 2 \log(\sqrt{2} \sigma) T\left(4,\frac{1}{2},\frac{1}{\sqrt{2} \sigma}\right) 
+ 2 T\left(5,\frac{1}{2},\frac{1}{\sqrt{2} \sigma}\right)
\biggr). \label{eq:in_gamma_3}
\end{align}

\paragraph{Expansions of $T(n,1/2,1/\sqrt{2}\sigma)$.} 
We need the values $T(3,1/2,1/\sqrt{2}\sigma), T(5,1/2,1/\sqrt{2}\sigma)$ as well as $T(5,1/2,1/\sqrt{2}\sigma)$. Under the assumption that $\sigma^2 > 1/2$, we can follow the arguments in \cite[(31), p.156; (36), p.157]{GED90} and \cite[p.144]{LUKE69}. For any $n \in \mathbb{N}$,
\[
T\left(n,\frac{1}{2},\frac{1}{\sqrt{2} \sigma} \right) = \frac{1}{2 \pi i} \oint_C \left(\frac{-1}{s+1}\right)^{n-1} \Gamma(-1/2 - s)(\sqrt{2}\sigma)^{-s} ds,
\]
where $i$ denotes the imaginary unit and $C$ is a contour given in \cite[(4), p.144]{LUKE69} which surrounds the poles at $s=-1$ and at $s= 1/2 + k$, for all $k \in \mathbb{N}$. The pole at $s=-1$ is of order $n-1$ and the residue for $n=3$ is
\begin{align*}
c_3:=& \frac{1}{2} \lim_{z\to -1} \frac{d^2}{dz^2} \frac{(z +1)^3}{(z+1)^3} \Gamma(-1/2 -z) (\sqrt{2}\sigma)^{-z} \\
=& \frac{\sigma}{\sqrt{2}} \Bigl(\frac{d^2}{dz^2} \Gamma\left(-\frac{1}{2} - z\right) \bigg|_{z=-1} \!\!\!\!\!\!
-2 \log(\sqrt{2}\sigma )  \frac{d}{dz} \Gamma\left(-\frac{1}{2} - z\right) \bigg|_{z=-1} \!\!\!\!\!\! +  \Gamma\left(\frac{1}{2}\right)   \log^2(\sqrt{2}\sigma ) \Bigr) \\
=& \frac{\sigma}{\sqrt{2}} \Bigl(\Gamma \left(\frac{1}{2} \right) \psi^2 \left(\frac{1}{2} \right) + \Gamma \left(\frac{1}{2} \right) \psi' \left(\frac{1}{2} \right) - 2\log(\sqrt{2}\sigma) \Gamma \left(\frac{1}{2} \right) \psi \left(\frac{1}{2} \right) +  \Gamma\left(\frac{1}{2}\right)   \log^2(\sqrt{2}\sigma )\Bigr) \\
=& \sigma \sqrt{\frac{\pi}{2}} \Bigl((2\log2 + \gamma + \log(\sqrt{2}\sigma))^2 + 3 \zeta(2) \Bigr) \approx \sigma \sqrt{\frac{\pi}{2}}((2.31 + \log \sigma)^2 + \pi^2 /2).
\end{align*}
For $n=4$ the residue is 
\begin{align*}
c_4 :=& - \frac{1}{6} \frac{d^3}{dz^3} \Gamma(-1/2 - z) (\sqrt{2}\sigma)^{-z} \Big|_{z=-1} \\
=& - \frac{\sqrt{2}\sigma}{6} \Bigl( \frac{d^3}{dz^3} \Gamma(-1/2 - z) \Big|_{z=-1} - 3 \log(\sqrt{2}\sigma)  \frac{d^2}{dz^2} \Gamma(-1/2 - z) \Big|_{z=-1} \\
&\quad\quad\quad\enspace+ 3 \log^2(\sqrt{2}\sigma)  \frac{d}{dz} \Gamma(-1/2 - z) \Big|_{z=-1} - \log^3(\sqrt{2}\sigma)  \Gamma(1/2)
\Bigr) \\
=& - \frac{\sqrt{2}\sigma}{6} \Bigl(\Gamma\left(\frac{1}{2}\right) \Bigl(\psi^3\left(\frac{1}{2} \right) + 3\psi\left(\frac{1}{2} \right) \psi'\left(\frac{1}{2} \right) + \psi''\left(\frac{1}{2} \right) \Bigr) \\
&\quad\quad\quad\enspace- 3\log(\sqrt{2}\sigma) \Bigl(\Gamma \left(\frac{1}{2} \right) \psi^2 \left(\frac{1}{2} \right) + \Gamma \left(\frac{1}{2} \right) \psi' \left(\frac{1}{2} \right) \Bigr) \\
&\quad\quad\quad\enspace+ 3\log^2(\sqrt{2}\sigma)\Gamma \left(\frac{1}{2} \right) \psi \left(\frac{1}{2} \right) - \log^3(\sqrt{2}\sigma) \Gamma \left(\frac{1}{2}\right)\Bigr)\\
=& \frac{\sigma \sqrt{2 \pi}}{6} \Bigl((2\log2 + \gamma + \log(\sqrt{2}\sigma))^3 + 9\zeta(2)(2\log2 + \gamma + \log(\sqrt{2}\sigma)) + 14\zeta(3) \Bigr)\\
\approx & \frac{\sigma \sqrt{2 \pi}}{6} ((2.31 + \log \sigma)^3 + \frac{3 \pi^2}{2}(2.31 + \log \sigma) + 16.828).
\end{align*}
Finally, for $n=5$, the residue is
\begin{align*}
c_5:=&  \frac{1}{2} \frac{d^4}{dz^4} \Gamma(-1/2 - z) (\sqrt{2} \sigma)^{-z} \Big|_{z=-1} \\
=& \frac{\sqrt{2}\sigma}{24} \Bigl(  \frac{d^4}{dz^4} \Gamma(-1/2 - z) \Big|_{z=-1}   -  4 \log(\sqrt{2}\sigma) \frac{d^3}{dz^3} \Gamma(-1/2 - z) \Big|_{z=-1} \\
&\quad\quad\quad+ 6 \log^2(\sqrt{2}\sigma) \frac{d^2}{dz^2} \Gamma(-1/2 - z) \Big|_{z=-1} - 4 \log^3(\sqrt{2}\sigma) \frac{d}{dz} \Gamma(-1/2 - z) \Big|_{z=-1} \\
&\quad\quad\quad+ \log^4(\sqrt{2}\sigma) \Gamma(1/2)
\Bigr) \\
=& \frac{\sigma \sqrt{2\pi}}{24} \Bigl( (2\log2 + \gamma + \log(\sqrt{2}\sigma))^4 + 18 \zeta(2)(2\log2 + \gamma + \log(\sqrt{2}\sigma))^2 \\
&\quad\quad\quad+ 56 \zeta(3) (2\log2 + \gamma + \log(\sqrt{2}\sigma)) + 27 \zeta^2(2) + 90 \zeta(4) \Bigr) \\
\approx & \frac{\sigma \sqrt{2\pi}}{24} ((2.31 + \log \sigma)^4 + 3 \pi^2 (2.31 + \log \sigma)^2 + 67.312 \log \sigma + 155.49 + \frac{7 \pi^4}{4})
\end{align*}
The residues at $z= 1/2 + k$, $k \in \mathbb{N}$, are 
\[
\left( \frac{-1}{3/2 + k} \right)^{n-1} \frac{(-1)^k}{k!} (\sqrt{2}\sigma)^{-(1/2 + k)}.
\]
Hence, 
\begin{gather*}
T\left(3,\frac{1}{2},\frac{1}{\sqrt{2}\sigma}\right) = c_3 + \sum_{k=0}^\infty \frac{(-1)^k}{k!(3/2 + k)^2 (\sqrt{2}\sigma)^{k+1/2}},\\
T\left(4,\frac{1}{2},\frac{1}{\sqrt{2}\sigma}\right) =  c_4  - \sum_{k=0}^\infty \frac{(-1)^k}{k!(3/2 + k)^3 (\sqrt{2}\sigma)^{k+1/2}},  \\ 
T\left(5,\frac{1}{2},\frac{1}{\sqrt{2}\sigma}\right) =  c_5 + \sum_{k=0}^\infty \frac{(-1)^k}{k!(3/2 + k)^4 (\sqrt{2}\sigma)^{k+1/2}}.
\end{gather*}
A good approximation of these values can be attained by summing up to $10$. For this case we get
\begin{align*}
T\left(3,\frac{1}{2},\frac{1}{\sqrt{2}\sigma}\right) \approx c_3 +  0.374 \sigma^{-0.5} - 0.095 \sigma^{-1.5} + 0.017 \sigma^{-2.5} - 0.0024 \sigma^{-3.5}, \\
T\left(4,\frac{1}{2},\frac{1}{\sqrt{2}\sigma}\right) \approx  c_4  - 0.374 \sigma^{-0.5} + 0.095 \sigma^{-1.5} - 0.017 \sigma^{-2.5} + 0.0024 \sigma^{-3.5},  \\ 
T\left(5,\frac{1}{2},\frac{1}{\sqrt{2}\sigma}\right) \approx  c_5 + 0.374 \sigma^{-0.5} - 0.095 \sigma^{-1.5} + 0.017 \sigma^{-2.5} - 0.0024 \sigma^{-3.5}.
\end{align*}

\paragraph{Derivation of $E(\log |X|)$,  $E(\log^2|X|)$,  $\Var(\log|X|)$ and $E(|\log^3 |X||)$.}
Now, integration by substitution allows us to derive the expected value, the variance, and the absolute third moment. In the following, assume that $X$ is a zero mean Gaussian random variable with variance $\sigma^2 >0$. Then
\begin{align*}
E(\log |X| ) &= \sqrt{2/\pi \sigma^2} \int_0^\infty \log(x) e^{-x^2/2 \sigma^2} \, dx \\
&= \frac{2}{\sqrt{\pi}} \int_0^\infty \log(x) e^{-x^2} \, dx + \frac{2}{\sqrt{\pi}} \log(\sqrt{2} \sigma) \int_0^\infty e^{-x^2} \, dx \\
&= -\frac{\gamma + 2 \log 2}{2} + \log \sigma + \frac{1}{2} \log 2 \\
&= \log \sigma - \frac{\gamma  + \log 2}{2}  \approx \log \sigma -0.63. 
\end{align*}
Similarly, we can derive $E(\log^2 |X| )$. In detail,
\begin{align*}
E(\log^2 |X| ) &= \sqrt{2/\pi \sigma^2} \int_0^\infty \log^2(x) e^{-x^2/2 \sigma^2} \, dx \\
&=(2/\sqrt{\pi}) \int_0^\infty (\log(\sqrt{2} \sigma) + \log(x))^2 e^{-x^2} \, dx \\
&= (2/\sqrt{\pi}) \log^2(\sqrt{2} \sigma) \int_0^\infty e^{-x^2} \, dx + (4/\sqrt{\pi}) \log (\sqrt{2} \sigma) \int_0^\infty \log(x) e^{-x^2} \, dx \\
&\quad + (2/\sqrt{\pi}) \int_0^\infty \log^2 (x) e^{-x^2} \, dx \\
&= \log^2(\sqrt{2} \sigma) - \log(\sqrt{2} \sigma) (\gamma + 2\log 2) + \frac{((2\log 2 + \gamma)^2 + \pi^2 /2)}{4}.
\end{align*}
Combining these we find the variance of $\log|X|$ to be
\begin{align}
\Var(\log|X|) &= E(\log^2|X|) - E(\log|X|)^2 \notag \\
&= \log^2(\sqrt{2} \sigma) - \log(\sqrt{2} \sigma) (  \gamma + 2 \log 2) + \frac{((2\log 2 + \gamma)^2 + \pi^2 /2)}{4}  \notag \\
&\quad- \frac{(2 \log \sigma - \gamma - \log2)^2}{4} \notag \\
&= \pi^2/8, \label{eq:variance_log}
\end{align}
which is independent of $\sigma >0$.

To derive the third absolute moment of $\log |X|$, we expand a third-order polynomial below and we apply the earlier derived results on the incomplete $\Gamma$ functions.   
\begin{align*}
&E(|\log^3|X||)    = \sqrt{2/\pi \sigma^2} \int_0^\infty |\log^3(x)| e^{-x^2/2 \sigma^2} \, dx \\
&= \sqrt{2/\pi \sigma^2} \left(\int_1^\infty \log^3(x) e^{-x^2/2 \sigma^2} \, dx - \int_0^1 \log^3(x) e^{-x^2/2 \sigma^2} \, dx 
 \right)
\\
&= \frac{2}{\sqrt{\pi}} \left( \int_{1/\sqrt{2}\sigma}^\infty (\log(\sqrt{2} \sigma) + \log(x))^3 e^{-x^2} \, dx - \int_0^{1/\sqrt{2}\sigma} (\log(\sqrt{2} \sigma) + \log(x))^3 e^{-x^2} \, dx 
\right).
\end{align*}
The third-order polynomial inside the integrals is
\[
\log^3(\sqrt{2} \sigma) + 3 \log^2(\sqrt{2} \sigma) \log(x) 
+ 3 \log(\sqrt{2} \sigma) \log^2(x) + \log^3(x).
\]
We will now address each of the terms in turn. The first term corresponds to 
\[
\log^3(\sqrt{2} \sigma) \left( \frac{2}{\sqrt{\pi}} \int_{1/\sqrt{2}\sigma}^\infty  e^{-x^2} \, dx - \frac{2}{\sqrt{\pi}} \int_0^{1/\sqrt{2}\sigma}  e^{-x^2} \, dx \right)  = \log^3(\sqrt{2} \sigma) (1   - 2 \text{erf}(1/\sqrt{2}\sigma)),
\]
where $\text{erf}$ is the error function.  
The second term corresponds to
\begin{align*}
&\frac{6}{\sqrt{\pi}} \log^2(\sqrt{2} \sigma) \left( 
 \int_{1/\sqrt{2}\sigma}^\infty \log(x)  e^{-x^2} \, dx -  \int_0^{1/\sqrt{2}\sigma} \log(x) e^{-x^2} \, dx \right) \\
&= \frac{6}{\sqrt{\pi}} \log^2(\sqrt{2} \sigma) \left(\frac{d}{ds}\Gamma(s,1/\sqrt{2}\sigma)\Big|_{s=1/2}  -
\frac{d}{ds}\gamma(s,1/\sqrt{2}\sigma)\Big|_{s=1/2}
\right) \\
&= \frac{6}{\sqrt{\pi}} \log^2(\sqrt{2} \sigma) \left(2 \frac{d}{ds}\Gamma(s,1/\sqrt{2}\sigma)\Big|_{s=1/2}  -
\frac{d}{ds}\Gamma(s)\Big|_{s=1/2}
\right).
\end{align*}
The third term is 
\begin{align*}
&\frac{6}{\sqrt{\pi}} \log(\sqrt{2} \sigma) \left( 
 \int_{1/\sqrt{2}\sigma}^\infty \log^2(x)  e^{-x^2} \, dx -  \int_0^{1/\sqrt{2}\sigma} \log^2(x) e^{-x^2} \, dx \right) \\
 &=\frac{6}{\sqrt{\pi}} \log(\sqrt{2} \sigma) \left( 
2 \frac{d^2}{ds^2}\Gamma(s,1/\sqrt{2}\sigma)\Big|_{s=1/2}  -
\frac{d^2}{ds^2}\Gamma(s)\Big|_{s=1/2}
\right).
\end{align*}
Similarly, the last term is equal to
\begin{align*}
\frac{2}{\sqrt{\pi}} \left(
2 \frac{d^3}{ds^3}\Gamma(s,1/\sqrt{2}\sigma)\Big|_{s=1/2}  -
\frac{d^3}{ds^3}\Gamma(s)\Big|_{s=1/2}
\right).
\end{align*}
Combining these, we find that under the assumption that $\sigma^2 > 1/2$, $\rho = E(|\log^3|X||)$  is equal to
\begin{align} 
&\log^3(\sqrt{2} \sigma) (1   - 2 \text{erf}(1/\sqrt{2}\sigma)) 
+\frac{6}{\sqrt{\pi}} \log^2(\sqrt{2} \sigma) \left(2 \frac{d}{ds}\Gamma(s,1/\sqrt{2}\sigma)\Big|_{s=1/2}  -
\frac{d}{ds}\Gamma(s)\Big|_{s=1/2}
\right) \notag \\
&+ \frac{6}{\sqrt{\pi}} \log(\sqrt{2} \sigma) \left( 
2 \frac{d^2}{ds^2}\Gamma(s,1/\sqrt{2}\sigma)\Big|_{s=1/2}  -
\frac{d^2}{ds^2}\Gamma(s)\Big|_{s=1/2}
\right)  \notag \\
&+ \frac{2}{\sqrt{\pi}} \left(
2 \frac{d^3}{ds^3}\Gamma(s,1/\sqrt{2}\sigma)\Big|_{s=1/2}  -
\frac{d^3}{ds^3}\Gamma(s)\Big|_{s=1/2}
\right). \label{eq:rho_value}
\end{align}
The derivatives of the $\Gamma$ function and incomplete $\Gamma$ function are given in Equations \eqref{eq:gamma_1}, \eqref{eq:gamma_2}, \eqref{eq:gamma_3}, \eqref{eq:in_gamma_1},\eqref{eq:in_gamma_2} and \eqref{eq:in_gamma_3}.

\paragraph{A bound on $\rho$.} \label{pg:bnd_on_rho} The above approach does not work  when $\sigma^2 \leq 1/2$ and we need an alternative approach for that. Even when the above approach works it might be convenient to have a simple expression that bounds $\rho$.
By an application of H\"older's inequality we get such a simple bound that holds for any $\sigma >0$. The bound is based on the $4$'th moment of $\log|X|$ which is
\begin{align*}
E(\log^4|X|) &= \sqrt{\frac{2}{\pi \sigma^2}} \int_0^\infty \log^4(x) e^{-x^2/2\sigma^2} dx \\
&= \frac{2}{\sqrt{\pi}} \int_0^\infty \log^4(\sqrt{2}\sigma x) e^{-x^2} \, dx \\
&= \log^4(\sqrt{2}\sigma) - 2\log^3(\sqrt{2}\sigma)(\gamma + 2\log2) + \frac{3}{2}\log^2(\sqrt{2}\sigma)((2\log2+\gamma)^2 + \pi^2/2)\\
&\quad\quad - \frac{1}{2}\log(\sqrt{2}\sigma) ((2\log2 + \gamma)^3 + \frac{3\pi^2}{2}(2\log2+\gamma) + 14\zeta(3)) \\
&\quad\quad + \frac{1}{16}((2\log2+\gamma)^4 + 3\pi^2(2\log2+\gamma)^2 + 56(2\log2 + \gamma)\zeta(3) + 7\pi^4/4).\\
\end{align*}
and
\begin{equation} \label{bnd:rho}
\rho = E(\log^3|X|) \leq (E(\log^4|X|))^{3/4}.
\end{equation}

\subsection{Independence of $S$ and $|X|$} \label{sec:ind_S_absX}
It is well known that the sign $S$ of a centered normal distributed random variable with variance $\sigma^2 > 0$ and its absolute value $|X|$ are independent. One way to verify this is to 
recall that $S$ and $|X|$ are independent if for all $a,b\in\mathbb{R}$, $\Pr(S \leq a, |X| \leq b) = \Pr(S \leq a) \Pr(|X| \leq b)$. It is easy to verify this: for any $b \in \mathbb{R}$ and $a < -1$
\[
\Pr(S_i \leq a, |X_i| \leq b)  = 0
= \Pr(S_i \leq a) \Pr(|X_i| \leq b)
\]
and for $a \geq 1$,
\[
\Pr(S_i \leq a, |X_i| \leq b)  = \Pr(|X_i| \leq b)=
\Pr(S_i \leq a) \Pr(|X_i| \leq b).
\]
Finally, for $-1 \leq a < 1$, and $b \geq 0$ ($b < 0$ is trivial), we have due to the symmetry of $X$ that
\begin{align*}
&\Pr(S_i \leq a) \Pr(|X_i| \leq b) = \Pr(|X_i| \leq b)/ 2  \\
&=  \Pr(-b \leq X_i \leq b)/2 = \Pr(0 \leq X_i \leq b) 
= \Pr(-b \leq X_i \leq 0)   \\
&= \Pr(X_i < 0, -X_i \leq b) 
= \Pr(X_i < 0, |X_i| \leq b) = 
\Pr(S_i \leq a, |X_i| \leq b).
\end{align*}

\section{Further Results on GPs and DGPs} \label{app:DGPs}

\subsection{Representing a GP with Quadratic Kernel by a Multivariate Normal RV} \label{app:rep_GP}
GPs can often be written as a (potentially infinite) linear combination of normally distributed random variables. That is the Karhunen-Lo\`eve expansion which is based on an eigendecomposition of the kernel-integral operator corresponding to the covariance function (Mercer's theorem). For polynomial kernels a more direct approach is possible. We demonstrate this approach here. We start with a simple 2-dimensional example, for which the key steps are transparent, before approaching the more abstract general case.

\paragraph{2-Dimensional feature map.}
Consider the following kernel function on $\mathbb{R}$,
\[
k_0(x,y) = x^2y^2 + xy = \begin{pmatrix}
x^2 \\
x 
\end{pmatrix}^\top \begin{pmatrix}
y^2 \\
y  
\end{pmatrix} = \psi(x)^\top \psi(y),
\]
where we will represent $\psi(x)$ as $\psi(x) = (\psi_1(x), \psi_2(x))^\top.$
The RKHS $\cH$ corresponding to $k_0$ is 2-dimensional since $x^2$ and $x$  are linearly independent.

We start by taking a closer look at $\cH_0$. Similarly as for the linear kernel in Section \ref{sec:MainSec}, we can write the RKHS in the form
\begin{align*}
\cH_0 &= \{h(x) = \alpha_1 \psi(-1)^\top\psi(x) + \alpha_2 \psi(-1/2)^\top \psi(x) : \alpha_1,\alpha_2 \in \mathbb{R}  \} \\
&= \{h(x) = \alpha_1 (x^2 - x) + \alpha_2 (\frac{1}{4} x^2 - \frac{1}{2} x)   : \alpha_1,\alpha_2 \in \mathbb{R}  \},
\end{align*}
since $\psi(-1)^\top\psi(x)$ and $\psi(-1/2)^\top \psi(x)$ are linearly independent and $\cH_0$ is 2-dimensional. Let $g_0$ be the GP corresponding to kernel $k_0$ then $g_0$ attains values in $\cH_0$ and there are random variables $\alpha_{1,\omega}$ and $\alpha_{2,\omega}$ such that for all $x\in \mathbb{R}$,
\[
g_0(x) = \alpha_{1,\omega} (x^2 - x) + \alpha_{2,\omega} (\frac{1}{4} x^2 - \frac{1}{2} x).
\]
Consider $x_1= -1$ then $g_0(-1) = 2 \alpha_{1,\omega} + (3/4) \alpha_{2, \omega}$. 
Furthermore, for $x_2=-1/2$ we find that $g_0(-1/2) = (3/4) \alpha_{1,\omega} + (5/16) \alpha_{2, \omega}.$ Let $C$ be a covariance matrix of a gaussian vector $(g_0(x_1), g_0(x_2))$, and is given by
$$
C = \begin{pmatrix}
k(x_1, x_1) & k(x_1, x_2)\\
k(x_2, x_1) & k(x_2, x_2) \end{pmatrix} = 
\begin{pmatrix}
    2 & 3/4\\
    3/4 & 5/16,
\end{pmatrix}
$$
since $k(-1, -1) = 1+1 = 2$, $k(-1, -1/2) = 3/4$, $k(-1/2, -1/2) = 5/16.$
Therefore, we get
$$
\begin{pmatrix}
g_0(-1)\\
g_0(-1/2)
\end{pmatrix}=
C \begin{pmatrix}
    \alpha_{1, \omega}\\
    \alpha_{2, \omega}
\end{pmatrix} = 
\begin{pmatrix}
    2 & 3/4\\
    3/4 & 5/16
\end{pmatrix}
\begin{pmatrix}
    \alpha_{1, \omega}\\
    \alpha_{2, \omega}
\end{pmatrix}.
$$
The covariance between $\alpha_{1,\omega}$ and $\alpha_{2,\omega}$ is given by
\begin{align*}
Cov \biggl(
\begin{pmatrix}
    \alpha_{1,\omega} \\ 
    \alpha_{2,\omega}
\end{pmatrix} \biggr) &= 
C ^{-1}  Cov \biggl(
\begin{pmatrix}
    g_0(-1) \\ 
    g_0(-1/2)
\end{pmatrix} \biggr) 
C ^{-1}\\
& = \begin{pmatrix}
5 & -12\\
-12 & 32
\end{pmatrix} 
\begin{pmatrix}
    2 & 3/4\\
    3/4 & 5/16
\end{pmatrix}
\begin{pmatrix}
5 & -12\\
-12 & 32
\end{pmatrix} \\
&=
\begin{pmatrix}
    5 & -12\\
    -12 & 32
\end{pmatrix} = C^{-1}.
\end{align*}
Hence, we get

\[
E(\alpha_{1,\omega} \alpha_{2,\omega}) = -12,
\]
and
\[
\begin{pmatrix}
    \alpha_{1,\omega} \\ 
    \alpha_{2,\omega}
\end{pmatrix}  \sim N\biggl(0, 
\begin{pmatrix}
5 & -12\\
-12 & 32
\end{pmatrix} \biggr).
\]
Rearranging the terms gives 
\begin{align*}
g_0(x) &= x^2(\alpha_{1,\omega}  + (1/4) \alpha_{2,\omega}) + x(\alpha_{1,\omega}  + (1/2) \alpha_{2,\omega})\\
&=  \psi_1(x)(\alpha_{1, \omega} \psi_1(x_1) + \alpha_{2, \omega} \psi_1(x_2)) + \psi_2(x)(\alpha_{1, \omega} \psi_2(x_1) + \alpha_{2, \omega} \psi_2(x_2))\\
&= \begin{pmatrix}
Y_1 \\
Y_2 
\end{pmatrix}^\top \psi(x),
\end{align*}
where
$$
\begin{pmatrix}
    Y_1 \\
    Y_2 
\end{pmatrix} = 
\begin{pmatrix}
    \psi_1(x_1) & \psi_1(x_2)\\
    \psi_2(x_1) & \psi_2(x_2)
\end{pmatrix}
\begin{pmatrix}
    \alpha_1 \\
    \alpha_2
\end{pmatrix}.
$$
It is easy to see that
$$
\begin{pmatrix}
    \psi_1(x_1) & \psi_1(x_2)\\
    \psi_2(x_1) & \psi_2(x_2)
\end{pmatrix} ^ \top \begin{pmatrix}
    \psi_1(x_1) & \psi_1(x_2)\\
    \psi_2(x_1) & \psi_2(x_2)
\end{pmatrix} = C.
$$
In the following, let 
\[B = \begin{pmatrix}
    \psi_1(x_1) & \psi_1(x_2)\\
    \psi_2(x_1) & \psi_2(x_2)
\end{pmatrix} \quad \text{ and note that } \quad \begin{pmatrix}
    Y_1 \\
    Y_2 
\end{pmatrix} =  B \begin{pmatrix}
    \alpha_1 \\
    \alpha_2
\end{pmatrix}.
\]
The covariance between $Y_1$ and $Y_2$ is 
$$
Cov \biggl(
\begin{pmatrix}
    Y_1 \\ 
    Y_2
\end{pmatrix} \biggr) = B \, Cov \biggl(
\begin{pmatrix}
    \alpha_{1,\omega} \\ 
    \alpha_{2,\omega}
\end{pmatrix} \biggr) B^\top.
$$
Following that, we multiply both sides by $B$ on the right, which gives
$$
Cov \biggl(
\begin{pmatrix}
    Y_1 \\ 
    Y_2
\end{pmatrix} \biggr) B = BC^{-1}B^\top B = B.
$$
Therefore,
$$
Cov \biggl(
\begin{pmatrix}
    Y_1 \\ 
    Y_2
\end{pmatrix} \biggr) = I, 
$$
and 
\[
\begin{pmatrix}
    Y_1 \\
    Y_2 
\end{pmatrix} \sim N\biggl(0, 
\begin{pmatrix}
 1 & 0\\
 0 & 1
\end{pmatrix} \biggr).
\]

\paragraph{The general case.} \label{app:polynomialKernel}
We  consider now the polynomial kernel function. The argument below generalizes, however, right away to any other  kernel function which has a finite dimensional feature vector. Denote the parameters of the polynomial kernel function 
with integer parameter $d \geq 1$, and real valued $c > 0$ on $\mathbb{R}$,
\[
k(x,y) = (xy + c)^d 
 = \phi(x)^\top \phi(y),
\]
where 
\[
\phi(x) = \left(x^d,  \binom{d}{1}^{1/2} \!\! x^{d-1} c^{1/2}, \binom{d}{2}^{1/2} \!\! x^{d-2} (c^2)^{1/2}, \ldots, c^{d/2} \right)^\top\!\!.
\]
The functions $x^d$, $x^{d-1} c^{1/2}, \ldots, c^{d/2}$ are linearly independent and \cite[Ex 3.7]{PAUL16} shows that these functions all lie in $\cH$. Hence, $\cH$ is at least $d+1$-dimensional. In fact, it follows from \cite[Thm 2.10]{PAUL16} that $\cH$ is $d+1$-dimensional. 
This implies that there are $x_1,\ldots, x_{d+1} \in \mathbb{R}$ such that
\[
\cH = \Bigl\{\sum_{i=1}^{d+1} \alpha_i k(x_i,\cdot) : \alpha_i \in \mathbb{R}, i \leq d+1 \Bigr\}.
\]
Let $g$ be a GP with kernel $k$ then $g$ attains values in $\cH$ and 
\[
g(x) = \sum_{i=1}^{d+1} \alpha_{i,\omega} k(x_i,x)
\]
for $d+1$ stochastic real valued coefficients $\alpha_{1,\omega},\ldots, \alpha_{d+1,\omega}$. In particular,
\[
\begin{pmatrix}
    g(x_1) \\
    \vdots \\
    g(x_{d+1})
\end{pmatrix} = C \pmb{\alpha}_\omega \quad \text{ and } \quad C^{-1} \begin{pmatrix}
    g(x_1) \\
    \vdots \\
    g(x_{d+1})
\end{pmatrix} = \pmb{\alpha}_\omega,
\]
where $\pmb{\alpha}_\omega = (\alpha_{1,\omega},\ldots, \alpha_{d+1,\omega})^\top$ and $C = (k(x_i,x_j))_{i,j\leq d+1 }$. That $C$ is invertible can be seen in the following way: the functions $k(x_1,\cdot), \ldots, k(x_{d+1}, \cdot)$ are linearly independent since they span the $d+1$-dimensional space $\cH$. In particular, for any $h$ there exists $a_1,\ldots, a_{d+1}$ such that $h = \sum_{i=1}^{d+1} a_i k(x_i,\cdot)$ and 
\[
\|h\|^2 = \pmb{a}^\top C \pmb{a},
\]
where $\pmb{a} = (a_1,\ldots, a_{d+1})^\top$. If $C$ would not be of full rank then there would exist an eigenvector $\pmb{e}$ of $C$ with eigenvalue $0$. The function $h$ corresponding to $\pmb{e}$ would not be the constant $0$ function since $\pmb{e}$ would not be zero. However, in this case
\[
0 \not = \|h\|^2 = \pmb{e}^\top C \pmb{e} =0,
\]
and $C$ has to be of full rank.

 Because $\pmb{\alpha}_\omega$ is a linear transformation of a zero mean Gaussian vector it follows that $\pmb{\alpha}_\omega$ is also a zero mean Gaussian vector. The matrix $C$ is the covariance matrix of the $d+1$-dimensional Gaussian vector $(g(x_1),\ldots, g(x_{d+1}))^\top$ and 
 \[
 Cov(\pmb{\alpha_\omega}) = C^{-1} C C^{-1} = C^{-1}.
 \]
Writing the Gaussian process in terms of the feature map $\phi$,
\[
g(x) = \sum_{i=1}^{d+1} \alpha_{i,\omega} \phi(x_i)^\top \phi(x) 
\]
and denoting the different entries of $\phi(x)$ by $\phi_1(x),\ldots,\phi_{d+1}(x)$, we  define $d+1$ zero mean Gaussian random variables $Y_1,\ldots, Y_{d+1}$ through 
\[
\sum_{i=1}^{d+1} \alpha_{i,\omega} \phi(x_i) 
= \underbrace{\begin{pmatrix}
\phi(x_1) & \ldots & \phi(x_{d+1}
\end{pmatrix}}_{B} 
\pmb{\alpha}_\omega
=
\begin{pmatrix}
Y_1 \\
\vdots \\
Y_{d+1}
\end{pmatrix}
\]
Note that $g(x) = Y^\top \phi(x)$. The random vector $Y = (Y_1,\ldots, Y_{d+1})^\top$ has covariance
$$
Cov \begin{pmatrix}
    Y_1 \\
    \vdots\\
    Y_{d+1}
\end{pmatrix} = 
B Cov(\pmb{\alpha}_\omega)B^\top.
$$
Multiplying  by $B$ on the right yields
\begin{align*}
Cov \begin{pmatrix}
    Y_1 \\
    \vdots\\
    Y_{d+1}
\end{pmatrix} B =  B Cov(\pmb{\alpha}_\omega) B^\top B 
 = B,
\end{align*}
since $B^\top B = C$. The matrix $B$ is invertible {\color{red}...} and multiplying the above from the right by $B^{-1}$ leads us to
$$
Cov \begin{pmatrix}
    Y_1 \\
    \vdots\\
    Y_{d+1}
\end{pmatrix} = I,
\quad \text{ and } \quad
\begin{pmatrix}
    Y_1\\
    \vdots\\
    Y_{d+1}
\end{pmatrix} \sim N(0, I).
$$


\subsection{A Berry-Esseen Bound for the case that $d_1 = \ldots = d_{\ell-1} = 2$} \label{app:BE_bnd_d2}
Before considering a general case, we first examine the approximation of DGPs, where the successive layers use a kernel $k_i(x,y) = (xy)^{d_i}$ with $d_i = 2$. We will also define $d^\downarrow_i = \sum_{j=0}^{i-1}2$ for all $i = 1, \dots, \ell-1$. As it was shown above, the GP $g_1$ can be represented as
$$
g_1(x) = \begin{pmatrix}
    Z_1\\
    \vdots\\
    Z_{d+1}
\end{pmatrix}^\top
\begin{pmatrix}
    \phi_1(x)\\
    \vdots\\
    \phi_{d+1}(x)
\end{pmatrix},
$$
where $(Z_1,\ldots, Z_{d+1})^\top \sim N(0, I)$. Also, let
$Y_i \sim N(0, \sigma_i^2)$ be i.i.d, and independent of $Z_1, \dots\ Z_{d+1}$, and such that $g_i(x) = Y_i x^2$ for all $2 \leq i \leq \ell$. Then the DGP can be written as

\begin{align*}
g_\ell \circ \dots \circ g_1(x) &= Y_\ell (Y_{\ell-1})^{d^\downarrow_1} (Y_{\ell -2})^{d^\downarrow_2} \times \dots \times (Y_{2})^{d^\downarrow_{\ell-2}} \Bigl(\sum_{i=1}^{d+1}Z_i \phi_i (x) \Bigr)^{d^\downarrow_{\ell-1}} \\
&= Y_\ell Y_{\ell-1}^{2} Y_{\ell -2}^{4} \times \dots \times Y_{2}^{2(\ell-2)} \Bigl(\sum_{i=1}^{d+1}Z_i \phi_i (x) \Bigr)^{2(\ell-1)}.
\end{align*}
Taking the logarithm of the absolute values of the product of $Y$-terms yields $2(\ell-2)\log|Y_2| + \dots + 2\log|Y_{\ell-1}| + \log|Y_\ell| = \sum_{j=2}^{\ell} c_j \log|Y_j|,$ where $c_j = 2(\ell-j)$ for $j = 2, \dots, \ell-1$ and $c_\ell = 1.$ We then want to apply the Berry-Esseen Theorem, and in order to do this we first define $\sigma_{i, \log} = c_i^2 \Var(\log|Y_i|)$ 
and $\rho_{i, \log} = c_i^3 E(|\log|Y_i||^3)$ for all $i=2,\dots, \ell$. Furthermore, to derive the expression for the Berry-Esseen bound, we first have


$$\Bigl(\sum_{i=2}^n \sigma_{i, \log}^2\Bigr)^{-3/2} \sum_{i=2}^n \rho_{i,\log} = \frac{\sum_{i=2}^\ell c_i^3 E(|\log|Y_i||^3)}{(\sum_{i=2}^\ell c_i^2)^{3/2} (\Var(\log|Y_i|))^{3/2}},$$
where the sums of the coefficients $\sum_{i=2}^{\ell-1} c_i^2$ and $\sum_{i=2}^{\ell-1} c_i^3$ can be found as  
\begin{align*}
\sum_{i=2}^{\ell-1} c_i^2 = 4\sum_{i=2}^{\ell-1}(\ell-i+1)^2 = 4\sum_{i=1}^{\ell-2}(\ell-i)^2 = 4 \sum_{i=1}^{\ell-1}i^2 -4 = \frac{2\ell(\ell-1)(2\ell-1)}{3} - 4,
\end{align*}

\begin{align*}
\sum_{i=2}^{\ell-1}c_i^3 &= 8\sum_{i=2}^{\ell-1} (\ell-i+1)^3 = 8\sum_{i=1}^{\ell-2}(\ell-i)^3\\ 
&= 8\Bigl(-\frac{1}{2}\ell^3(\ell-1) + \frac{1}{2}\ell^2(\ell-1)(2\ell-1) - \sum_{i=1}^{\ell-1}i^3 -1 \Bigr)\\
&= 8\Bigl(-\frac{1}{2}\ell^3(\ell-1) + \frac{1}{2}\ell^2(\ell-1)(2\ell-1) - \frac{\ell^2(\ell-1)^2}{4} -1 \Bigr) \\
&= 2\ell^2(\ell-1)^2 - 8.
\end{align*}
For later use we also find
\[
\sum_{i=2}^{\ell}c_i=\sum_{i=2}^{\ell-1}c_i +1 =2\sum_{i=2}^{\ell-1}(\ell -i +1)+1=2\sum_{i=2}^{\ell-2}(\ell -i) +1= \ell(\ell-1)-2+1 = \ell(\ell-1)-1
\]
Hence, we get $\sum_{i=2}^\ell c_i^2 = \frac{2\ell(\ell-1)(2\ell-1)}{3} -3$, and $\sum_{i=2}^\ell c_i^3 = 2\ell^2(\ell-1)^2 -7$, and this gives us the expressions for the bound, the mean and the variance of $Y$
\begin{gather}
\begin{aligned}
&0.56 \Bigl(\sum_{i=2}^\ell \sigma^2_{i,\log} \Bigr)^{-3/2} \sum_{i=2}^\ell \rho_{i,\log} = 0.56 \frac{E(|\log|Y_1||^3)}{(\Var(\log|Y_1|))^{3/2}} \frac{(2\ell^2(\ell-1)^2-7)3^{3/2}}{(2\ell(\ell-1)(2\ell-1)-9)^{3/2}} \\
&\leq 3 \ell^{-1/2} \frac{E(|\log|Y_1||^3)}{(\Var(\log|Y_1|))^{3/2}},
\end{aligned} \label{eq:gather_BE}\\
\begin{aligned}
\sum_{i=2}^\ell c_i E(\log|Y_i|) = \sum_{i=2}^\ell c_i E(\log|Y_i|) &\approx (\ell(\ell-1) -1)(\log \sigma -0.63)\\
&= ((\ell-1)^2+\ell)(\log\sigma -0.63), 
\end{aligned} \label{eq:gather_Mean} \\
\begin{aligned}
\Var(\sum_{i-2}^\ell c_i^2 \log|Y_i|) = \sum_{i-2}^\ell c_i^2 \Var(\log|Y_i|) &= \Bigl(\frac{2\ell(\ell-1)(2\ell-1)}{3}-3\Bigr)\Var(\log |Y_1|) \\
&= \frac{\pi^2}{8}\Bigl(\frac{2\ell(\ell-1)(2\ell-1)}{3}-3\Bigr). \label{eq:gather_Var}
\end{aligned}
\end{gather}

The Berry-Esseen Theorem guarantees that 
\begin{align*}
&\sup_{x\in \mathbb{R}} |\Pr((\sigma^2_{2,\log} + \dots + \sigma^2_{\ell,\log})^{-1/2} \sum_{j=2}^\ell (c_j \log |Y_j| - c_jE(\log|Y_j|))\leq x) - \Phi(x)| \notag \\ 
&\leq 0.56 
\Bigl(\sum_{i=1}^n \sigma_{i,\log}^2\Bigr)^{-3/2} \sum_{i=1}^n \rho_{i,\log}.
\end{align*}
Similarly, to the derivation in Section \ref{sec:DGPApprox} by substitution, 
$$
\sup_{x\in \mathbb{R}} |\Pr\Bigl( |Y_\ell| \prod_{i=2}^{\ell-1} |Y_i|^{2(\ell-i)} \leq x\Bigr)
- \Pr(e^Y \leq x)| = \sup_{x\in \mathbb{R}} |\sum_{j=2}^\ell c_j \log |Y_j| - \Pr(Y \leq x)|,
$$
where $Y \sim N(\sum_{i=2}^\ell c_i E(\log|Y_i|), \sum_{i=2}^\ell \sigma^2_{i,\log})$ and the last term is upper bounded by the bound given in \eqref{eq:gather_BE}. Note that $Y_\ell \prod_{i=2}^{\ell-1} Y_i^{2(\ell-i)} = S_\ell \prod_{i\in I} S_i |Y_\ell| \prod_{i=2}^{\ell-1} |Y_i|^{2(\ell-i)}$, where we can write $S = S_\ell \prod_{i\in I} S_i$, since $S$ has the same distribution as the product on the right hand side, and attains values $-1$ and $1$ with probability $1/2$.

\subsection{Incorporating $(g_1(x))^{c_1}$ in the Approximation} \label{sec:conditioning}
In Section \ref{sec:DGPApprox} we derive an approximation that leads us to a bound on
\[
\sup_{y\in \mathbb{R}} \bigl|\Pr\Bigl(S \prod_{i=2}^\ell Y_i \leq y\Bigr) - \Pr(S e^Y \leq y)\bigr|.
\]
The involved random variables are specified in Section \ref{sec:DGPApprox}. We also have a random variable $(g_1(x))^{c_1}$ that we like to incorporate on both sides. In detail, we want a bound on 
\[
\sup_{y\in \mathbb{R}} \bigl|\Pr\Bigl(S (g_1(x))^{c_1} \prod_{i=2}^\ell Y_i \leq y\Bigr) - \Pr(S (g_1(x))^{c_1} e^Y \leq y)\bigr|.
\]
We can attain such a bound by using a conditional expectation argument.
Note that, due to the towering rule of conditional expectations,
\begin{align*}
&\Pr\Bigr(S (g_1(x))^{c_1} \sum_{j=2}^\ell c_j \log |Y_j| \leq y \Bigr)  =  E\Bigl( \pmb{1} \Bigl\{S (g_1(x))^{c_1} \sum_{j=2}^\ell c_j \log |Y_j|\leq y\Bigr\}\Bigr) \\
&= E\Bigl( E\Bigl( \pmb{1} \Bigl\{S (g_1(x))^{c_1} \sum_{j=2}^\ell c_j \log |Y_j|\leq y\Bigr\} \Big|  (g_1(x))^{c_1}\Bigr)\Bigr),
\end{align*}
where $\pmb{1}$ denotes the indicator function. Since the indicator function is real-valued, there exists a measurable function $h$ such that the conditional expectation is equal to $h((g_1(x))^{c_1})$ (see, for example, \cite[Eq. 10, p.220]{SHIR95}). In fact, one can observe that we can choose
\[
h(z) = E\Bigl( \pmb{1} \Bigl\{S z \sum_{j=2}^\ell c_j \log |Y_j|\leq y\Bigr\}\Bigr)
= \Pr\Bigl(Sz \sum_{j=2}^\ell c_j \log |Y_j|\leq y\Bigr).
\]
The same approach leads us to a function 
\[
\tilde h(z) = \Pr\Bigl(Sz e^Y \leq y\Bigr)
\]
and 
\[
\sup_{z\in \mathbb{R}} |h(z) - \tilde h(z)| 
=\sup_{y\in \mathbb{R}} \bigl|\Pr\Bigl(S \prod_{i=2}^\ell Y_i \leq y\Bigr) - \Pr(S e^Y \leq y)\bigr|.
\]
Hence,
\begin{align*}
&\sup_{y\in \mathbb{R}} \bigl|\Pr\Bigl(S (g_1(x))^{c_1} \prod_{i=2}^\ell Y_i \leq y\Bigr) - \Pr(S (g_1(x))^{c_1} e^Y \leq y)\bigr|  \\
&=  \sup_{y\in \mathbb{R}} \bigl|
E(h((g_1(x))^{c_1})) - E(\tilde h((g_1(x))^{c_1}))|  
\bigr|  \\
&\leq  \sup_{z\in \mathbb{R}} |h(z) - \tilde h(z)| \\
&= \sup_{y\in \mathbb{R}} \bigl|\Pr\Bigl(S \prod_{i=2}^\ell Y_i \leq y\Bigr) - \Pr(S e^Y \leq y)\bigr|
\end{align*}
and bound is not affected by the introduction of $(g_1(x))^{c_1}$.

\bibliographystyle{unsrt}  
\bibliography{WienerDeepGP_arxiv}

\begin{thebibliography}{10}

\bibitem{pmlr-v31-damianou13a}
A.~Damianou and N.~D. Lawrence.
\newblock Deep {G}aussian processes.
\newblock In {\em Proceedings of the Sixteenth International Conference on Artificial Intelligence and Statistics}, volume~31, 2013.

\bibitem{NEURIPS2022_8c420176}
M.~Giordano, K.~Ray, and J.~Schmidt-Hieber.
\newblock On the inability of gaussian process regression to optimally learn compositional functions.
\newblock In {\em Advances in Neural Information Processing Systems}, volume~35, 2022.

\bibitem{JMLR:v24:21-0556}
G.~Finocchio and J.~Schmidt-Hieber.
\newblock Posterior contraction for deep gaussian process priors.
\newblock {\em Journal of Machine Learning Research}, 24(66):1--49, 2023.

\bibitem{abraham2023deepgaussianprocesspriors}
K.~Abraham and N.~Deo.
\newblock Deep gaussian process priors for bayesian inference in nonlinear inverse problems, 2023.

\bibitem{castillo2024}
I.~Castillo and T.~Randrianarisoa.
\newblock Deep horseshoe gaussian processes, 2024.

\bibitem{pmlr-v33-duvenaud14}
D.~Duvenaud, O.~Rippel, R.~Adams, and Z.~Ghahramani.
\newblock {Avoiding pathologies in very deep networks}.
\newblock In {\em Proceedings of the Seventeenth International Conference on Artificial Intelligence and Statistics}, 2014.

\bibitem{DUN18}
M.~M. Dunlop, M.~A. Girolami, A.~M. Stuart, and A.~L. Teckentrup.
\newblock How deep are deep gaussian processes?
\newblock {\em Journal of Machine Learning Research}, 19(1), 2018.

\bibitem{kallenberg02}
O.~Kallenberg.
\newblock {\em Foundations of Modern Probability}.
\newblock Springer, third edition, 2021.

\bibitem{vanderVaart1998}
A.~W. van~der Vaart.
\newblock {\em Asymptotic Statistics}.
\newblock Cambridge University Press, 1998.

\bibitem{fremlin_v1}
D.H. Fremlin.
\newblock {\em Measure Theory, Volume 1}.
\newblock Torres Fremlin, 2000.

\bibitem{GED90}
K.O. Geddes, M.L. Glasser, R.A. Moore, and T.C. Scott.
\newblock Evaluation of classes of definite integrals involving elementary functions via differentiation of special functions.
\newblock {\em Applicable Algebra in Engineering, Communication and Computing}, 1, 1990.

\bibitem{LUKE69}
Y.~L. Luke.
\newblock {\em The special functions and their approximations}, volume~1.
\newblock London: Academic Press, 1969.

\bibitem{PAUL16}
V.~Paulsen and M.~Raghupathi.
\newblock {\em An Introduction to the Theory of Reproducing Kernel Hilbert Spaces}.
\newblock Cambridge University Press, 2016.

\bibitem{SHIR95}
A.~N. Shiryaev.
\newblock {\em Probability (2nd ed.)}.
\newblock Springer-Verlag, Berlin, Heidelberg, 1995.

\end{thebibliography}

\end{document}